\pdfoutput=1
\documentclass[a4paper,fleqn]{cas-dc}

\usepackage[numbers,sort&compress]{natbib}
\newcolumntype{L}[1]{>{\raggedright\arraybackslash}p{#1}}
\usepackage{url}
\usepackage{algorithm}
\usepackage{algpseudocode}

\graphicspath{{figures/}}

\newcommand{\dcbreak}{\nonumber\\&\quad}
\begin{document}

\shorttitle{ASPIRE: agentic identification of closed-form models}

\title[mode=title]{ASPIRE: An Agentic System-Identification Platform for Iterative Reasoning and Exploration}

\shortauthors{E. Derner, J. Kubal\'ik and R. Babu\v{s}ka}

\author[1]{Erik Derner}
\cormark[1]
\ead{erik.derner@cvut.cz}

\affiliation[1]{organization={Czech Institute of Informatics, Robotics and Cybernetics, Czech Technical University in Prague},
  addressline={Jugosl\'avsk\'ych partyz\'an\r{u} 1580/3},
  postcode={160 00},
  city={Prague 6},
  country={Czech Republic}}

\author[1]{Ji\v{r}\'i Kubal\'ik}
\ead{jiri.kubalik@cvut.cz}

\author[1,2]{Robert Babu\v{s}ka}
\ead{r.babuska@tudelft.nl}

\affiliation[2]{organization={Department of Cognitive Robotics, Delft University of Technology},
  addressline={Mekelweg 2},
  postcode={2628 CD},
  city={Delft},
  country={The Netherlands}}

\cortext[1]{Corresponding author}

\RenewDocumentCommand\printorcid{}{}

\ExplSyntaxOn
\cs_set:Npn \__first_footerline: { \group_begin: \small \sffamily \__short_authors: \group_end: }
\ExplSyntaxOff

\newcommand{\repositorylink}{\url{https://github.com/erik-derner/aspire-sysid}}

\begin{abstract}
System identification requires expert decisions on the identification method, incorporation of prior physical knowledge, and model validation. We propose ASPIRE, a framework for nonlinear system identification in which a large language model agent selects and executes the identification method. ASPIRE constructs closed-form, nonlinear static or dynamic models from measured data, optionally supplemented by a system description, the intended use of the model, and physical requirements expressed formally or in natural language. Each model includes its parameters, an estimation record, an interpretation of its terms, and an error evaluated on test data inaccessible to the agent. Formal physical requirements are verified independently. ASPIRE operates autonomously or interactively, allowing the agent to ask questions and the user to request further analyses and identification rounds. We evaluate ASPIRE on three measured systems: a mobile robot, a rotational system with dry friction, and a laboratory fermenter. Each system is tested with and without a description, in five runs per task. All 30 runs delivered valid models, at a cost of 0.96 to 3.38 US dollars per run. An assessment by the authors found that, for the 30 models recommended by the agent, one per run, the interpretation of the model and the documentation of its structure selection and parameter estimation fully satisfied 132 of 150 assessed items and partly satisfied 18. The most frequent discrepancy, found in nine models, concerned inaccurate or unsupported claims about model structure selection.
\end{abstract}


\maketitle

\section{Introduction}
\label{sec:intro}

In industrial applications, white-box models based on closed-form mathematical expressions are usually preferred to black-box models. Closed-form models relate inputs, states, and outputs through a small set of named parameters and, as such, are useful wherever a model must be interpreted, checked against physical knowledge, extrapolated beyond the range of the data, or embedded in model-based control. Obtaining such models from measured data requires several decisions by an expert: the identification method, the model class, how prior physical knowledge enters the estimation, the estimation criterion, and the validation procedure that matches the model's intended use. Techniques such as symbolic regression (SR) and sparse regression automate the search for model structure~\cite{brunton2016sindy,cranmer2023pysr}, and some SR tools allow for incorporating physical requirements and prior knowledge~\cite{kubalik2020gecco,cornelio2023aidescartes,corywright2024aihilbert,kubalik2026ansr,xiao2026pgsr}.

Agents based on large language models (LLMs) that write and execute code can choose the identification method: they can inspect the data, propose model structures informed by physical knowledge, implement a fitting procedure, or call an existing SR package and iterate. Recent systems couple LLM-proposed equation structures with numerical parameter fitting~\cite{shojaee2025llmsr,grayeli2024lasr}, give an LLM a fixed set of code-execution tools for equation discovery~\cite{xia2026srscientist}, or let an agent apply physics-based analysis tools to infer properties of the system and then configure existing SR and sparse-regression packages accordingly~\cite{yang2026kepleragent}. In system identification, ASIA~\cite{piga2026asia} delegates the choice of the model class, the training algorithm, and the hyperparameters to an autonomous coding agent.

Existing agentic equation-discovery methods have several limitations: they do not support the specification and verification of physics-based requirements, evaluation on test data inaccessible to the agent, and documentation of the identification procedure. Furthermore, each delivered model should include a record of its structure and parameter estimation, distinguishing estimated parameters, fixed constants, and parameters derived from other parameters. The record should contain sufficient information to check the reported identification procedure against the executed code and execution logs.
To the best of our knowledge, no published framework combines these features with agent-selected identification methods and a protocol for reporting variability across repeated runs.

We propose ASPIRE\footnote{ASPIRE stands for Agentic System-identification Platform for Iterative Reasoning and Exploration. It is unrelated to the robot-skill system of the same name~\cite{nvidia2026aspire}.}, a framework for system identification using LLM agents. ASPIRE constructs closed-form static or dynamic models from measured data, optionally supplemented by a system description, the model's intended use, and physical requirements expressed formally or in natural language. The agent selects and executes the identification method, which may involve physically motivated model structures, symbolic or sparse regression, or custom estimation code. Each candidate model is accompanied by its parameter values, estimation procedure, interpretation, limitations, and test error. The agent also recommends one of the candidates. ASPIRE supports autonomous and interactive operation; in the latter, the agent can request information from the user, who can also request analyses and further identification rounds. The framework consists of one or more LLM agents responsible for identification and a software harness that executes their commands, enforces computational budgets, checks and evaluates candidate models, and records the identification runs.

In summary, the main contributions of this paper are:

\begin{itemize}

\item We introduce ASPIRE, an agentic framework for system identification in which LLM agents autonomously select, implement, and execute identification methods to obtain nonlinear closed-form models from measured data. Unlike systems based on predefined modeling pipelines or restricted toolboxes, ASPIRE allows the agent to develop and compare alternative model structures and estimation procedures, including methods not specified in advance.

\item We integrate physical knowledge and user interaction into agentic system identification. Users can specify the intended application, physical requirements, and available knowledge about the system, while the agent can request clarifications, explain its results, and revise models based on user feedback. Formal physical requirements are checked independently of the agent, including properties such as monotonicity, equilibrium conditions, and physically consistent behavior under specified operating conditions, which cannot be assessed by predictive accuracy alone.

\item We develop an identification framework that separates autonomous model construction from independent model verification and evaluation. The framework prevents agents from accessing test data, verifies formal physical requirements, and retains executable models, parameter estimation records, and execution logs. This separation enables the identification procedure and its results to be examined and reproduced, despite the agent's freedom to choose its methods.

\item We demonstrate the framework on three experimentally measured systems: a mobile robot, a rotational system with dry friction, and a laboratory fermenter. Repeated experiments with and without system descriptions investigate the agent's ability to identify models from data alone and to incorporate prior physical knowledge. The results show that the agent can construct models with competitive predictive accuracy while satisfying physical requirements that are not enforced by conventional symbolic regression and sparse identification methods. The experiments also quantify run-to-run variability and computational cost.

\end{itemize}

The paper is organized as follows. Section~\ref{sec:related} reviews related work. Section~\ref{sec:problem} defines the identification problem. Section~\ref{sec:method} describes the ASPIRE framework. Section~\ref{sec:setup} describes the test systems, the compared methods, and the experimental design. Section~\ref{sec:results} presents the results, and Section~\ref{sec:discussion} discusses them. Section~\ref{sec:conclusion} concludes the paper.

\section{Related Work}
\label{sec:related}

\subsection{LLM-based and agentic equation discovery}
\label{sec:related-llm}

A prominent early example of LLM-in-the-loop equation discovery is FunSearch~\cite{romeraparedes2024funsearch}, which combines LLM-generated programs with automated evaluation in an evolutionary loop. AlphaEvolve~\cite{novikov2025alphaevolve} extends this paradigm from evolving individual functions to evolving programs comprising multiple functions and components. LLM-SR~\cite{shojaee2025llmsr} also builds on the FunSearch framework, adapting it to scientific equation discovery by combining LLM-proposed equation structures with numerical optimization of their constants.
Several refinements retain this fixed-pipeline character~\cite{grayeli2024lasr,du2024llm4ed,wang2025drsr,merler2024icsr,abhyankar2026llmaces}, including hybrids that embed the LLM as a guided variation operator inside genetic-programming search over dynamical-system equations~\cite{ziaeibideh2026llmode}. A complementary line uses LLMs to synthesize components of the evolutionary search itself, such as parent-selection operators~\cite{zhang2026parentselection}.
Recent agentic approaches give LLMs control over the computational steps of equation discovery. SR-Scientist~\cite{xia2026srscientist} provides an LLM with a fixed set of code-execution tools for data analysis and equation evaluation, which it uses to iteratively refine candidate equations. KeplerAgent~\cite{yang2026kepleragent} uses physics-based tools to infer properties of the system and configure symbolic-regression methods such as PySR and PySINDy. STRIDE~\cite{su2026stride} organizes equation discovery into modules with fixed roles and uses self-reflection to guide corrections.
In system identification, ASIA~\cite{piga2026asia} uses an LLM as an autonomous coding agent built on existing agentic platforms. Given a plain-language problem description, the agent chooses the model class, training algorithm, and hyperparameters. ASIA was tested on two system-identification benchmarks. Its authors identify implicit test leakage, reduced methodological transparency, and reduced reproducibility as limitations.
The environment in which an agent operates is examined in~\cite{he2026doe}, where production coding agents are treated as opaque model-discovery operators, and in a position paper on the design of agent environments~\cite{eurekagent2026}. Neither work addresses system identification with physical requirements and closed-form model outputs.

Table~\ref{tab:comparison} compares ASPIRE with the three related systems. As in ASIA, the agent of ASPIRE chooses its identification method, and it is not restricted to a fixed set of methods as in SR-Scientist and KeplerAgent. ASPIRE differs from all three systems in the role of physical knowledge and in the output it delivers. The user states physical requirements either formally or in natural language, and the harness verifies the formal requirements for every delivered model. Every model is a closed-form expression with a record of its parameters, estimation procedure, and interpretation, and the execution logs allow this record to be checked against the agent's actions. Moreover, the evaluation protocol keeps the test data inaccessible to the agent, which addresses the test leakage named by the authors of ASIA.
\begin{table*}[pos=tp]
\centering
\caption{Comparison of ASPIRE with three related agentic identification and equation-discovery systems. The entries for ASIA, SR-Scientist, and KeplerAgent are based on the cited papers.}
\label{tab:comparison}
\small
\begin{tabular}{L{0.13\textwidth}L{0.22\textwidth}L{0.17\textwidth}L{0.17\textwidth}L{0.19\textwidth}}
\toprule
Property & ASPIRE & ASIA~\cite{piga2026asia} & SR-Scientist~\cite{xia2026srscientist} & KeplerAgent~\cite{yang2026kepleragent} \\
\midrule
Identification method
& chosen and implemented by the agent; no predefined method required
& agent selects and implements model classes, training methods, and hyperparameters
& agent writes analysis code and searches for equations using predefined tools
& agent configures PySR and PySINDy using data analysis and inferred physical properties \\
\\
Delivered model
& closed-form expressions with parameter and estimation records
& models from agent-selected classes, including neural and hybrid models
& symbolic equation with fitted constants
& symbolic equations, including differential equations \\
\\
Physical knowledge of the user
& descriptions and requirements in natural language; formal requirements verified
& problem description, optional domain knowledge, and modeling suggestions
& no documented support for user-specified physical requirements
& physical properties inferred from data; no documented verification of user requirements \\
\\
Test data
& inaccessible to the agent; evaluated by the harness
& held-out tests; potential implicit leakage acknowledged
& separate test sets; agent isolation not specified
& test error used for stopping and selection across runs \\
\bottomrule
\end{tabular}
\end{table*}

\subsection{Physical constraints and prior knowledge}
\label{sec:related-constraints}

Physical knowledge was used to constrain and guide symbolic regression before the introduction of LLM-based equation discovery. Existing approaches include multi-objective SR with constraint samples~\cite{kubalik2020gecco,kubalik2021eswa} and search initialized with prior features~\cite{derner2021priors}. Other methods add constraint losses into neural SR~\cite{kubalik2023n4sr,kubalik2025en4sr}. These neural approaches have also been extended using asynchronous neuro-evolutionary search~\cite{kubalik2026ansr}.
AI-Descartes and AI-Hilbert incorporate background axioms through logical reasoning and algebraic geometry~\cite{cornelio2023aidescartes,corywright2024aihilbert}. PG-SR encodes prior knowledge as executable constraint programs~\cite{xiao2026pgsr}.
These methods use physical knowledge within predefined search procedures. ASPIRE supplies physical requirements to an agent that chooses its own identification procedure. The harness checks the formal requirements on each delivered model (Section~\ref{sec:task}).

\subsection{Autonomous machine-learning and research agents}
\label{sec:related-agents}

LLM-based agents also automate general machine-learning engineering. AIDE uses tree search over solution scripts~\cite{jiang2025aide}. Other examples include R\&D-Agent~\cite{rdagent2025}, DS-Agent~\cite{guo2024dsagent} and Kaggle-scale agents~\cite{grosnit2024agentk}. MLE-bench provides a benchmark for evaluating such agents~\cite{chan2025mlebench}.
Research-automation systems include the AI Scientist~\cite{lu2024aiscientist,yamada2025aiscientistv2}, Curie~\cite{kon2025curie}, CodeScientist~\cite{jansen2025codescientist} and data-to-paper~\cite{ifargan2025datatopaper}. SWE-agent examines the interface between a language model and a software environment~\cite{yang2024sweagent}.
These systems deliver arbitrary models or artifacts, and none of them requires a closed-form model or verifies physical requirements. Agentic LLM systems have also been surveyed in robotics, where deployed systems are graded by their degree of autonomy, goal-directed behavior, adaptability, and decision-making~\cite{raptis2025agenticness}.

\subsection{Evaluation conventions}
\label{sec:related-eval}

We report predictive accuracy and model complexity, as in SRBench~\cite{lacava2021srbench}. For dynamic systems, the identified model is simulated on test trajectories using only the measured inputs~\cite{nonlinearbenchmark,dascoli2024odeformer}.
Shared SR infrastructure such as SRToolkit~\cite{meznar2026srtoolkit} standardizes pipeline components to prevent implementation details from confounding method comparisons. ASPIRE addresses the evaluation of agents, for which the evaluator must be sealed against the system under test.
We do not run SR-Scientist, KeplerAgent or ASIA on the tasks of this paper. None of these systems accepts formal requirements, a specified evaluation mode and sealed test data as inputs; running them under the conditions of this paper would require adaptations of each system, and the results would then describe the adaptation rather than the published system. As reference points for the test error, we use two established tools for closed-form models with published default settings, PySINDy~\cite{brunton2016sindy,desilva2020pysindy} and PySR~\cite{cranmer2023pysr}. Like the agentic systems, these tools do not accept physical requirements; we therefore report their test scores and, separately, whether their models violate the formal requirements of the task (Section~\ref{sec:references}).

\section{Problem Definition}
\label{sec:problem}

We consider the identification of closed-form models of static and dynamic systems from measured data. The user specifies the available data, the required model form, and the evaluation procedure. Additional information, including a description of the system, its intended use, and physical requirements, may be provided, but is not mandatory. We first define the data and model representations supported by ASPIRE, followed by the assumptions and the identification objective.

The datasets are provided as tables with the columns \emph{name}, \emph{unit}, \emph{description}, \emph{roles}, and \emph{training data range}. A \emph{trajectory} is an ordered sequence of samples from one experiment, recorded with sampling period $T_s$. Each column may have one or more of the following roles: \emph{input}, an exogenous signal read from the data during simulation; \emph{state}, a component of the state vector computed by the model during simulation; \emph{output}, a measured signal used for model evaluation; and \emph{state derivative}, the sampled time derivative of a declared state, which may serve as a fitting target.

We denote the input vector by $u \in \mathbb{R}^{n_u}$, the state vector by $x \in \mathbb{R}^{n_x}$, and the measured output by $y \in \mathbb{R}$. The vectors $p$ and $c$ contain the parameters estimated from the data and the fixed constants, respectively. A hat denotes a model prediction or a parameter estimate.

\paragraph{Static models.}
A static model describes a memoryless relation between the current input and the output. The agent identifies an expression $f$ for the predicted output,
\begin{equation}
\hat{y} = f(u; p, c),
\label{eq:static}
\end{equation}
where $f$ is a scalar-valued function.

\paragraph{Dynamic models.}
A dynamic model describes the evolution of the system state. ASPIRE supports both continuous-time and discrete-time state-space representations. For continuous-time systems, the agent identifies the state-derivative function $g_c$,
\begin{equation}
\frac{\mathrm{d}x}{\mathrm{d}t} = g_c(x, u; p, c),
\label{eq:ct}
\end{equation}
where $t$ denotes continuous time. For discrete-time systems, the agent identifies the state-transition function $g_d$,
\begin{equation}
x(k+1) = g_d\bigl(x(k), u(k); p, c\bigr),
\label{eq:dt}
\end{equation}
where $k$ is the sampling instant. The functions $g_c$ and $g_d$ are vector-valued, with one expression for each state component. The measurement equation maps the state and input to the predicted output,
\begin{equation}
\hat{y} = h(x, u),
\label{eq:meas}
\end{equation}
where $h$ is given and may simply select one state component.

\paragraph{NARX models.}
A discrete-time task may also be formulated as a nonlinear autoregressive model with exogenous inputs (NARX)~\cite{leontaritis1985narx},
\begin{align}
\hat{y}(k+1) &= f_\mathrm{N}\bigl(y(k), \dots, y(k-n_a+1), \dcbreak u(k), \dots, u(k-n_b+1); p, c\bigr),
\label{eq:narx}
\end{align}
where $f_\mathrm{N}$ is the model expression, and $n_a$ and $n_b$ are the numbers of past outputs and inputs, respectively, specified by the task.

The NARX model is a special case of the discrete-time state-space model (\ref{eq:dt}) with measurement equation (\ref{eq:meas}). Its state vector is
$x(k) = [y(k), \dots, y(k-n_a+1)]^\top$, with components $x_1, \dots, x_{n_a}$. The first component of $g_d$ is the unknown function $f_\mathrm{N}$, whereas the remaining components are defined by the known shift equations
$x_i(k+1) = x_{i-1}(k)$, $i = 2, \dots, n_a$.
Thus, $f_\mathrm{N}$ is the only expression to be identified, and $h$ selects the first state component $x_1$. The past inputs $u(k-1), \dots, u(k-n_b+1)$ are read from the data in the same way as the current input $u(k)$.

The distinction between prediction and simulation is relevant. In one-step-ahead prediction, the past outputs in (\ref{eq:narx}) are taken from the measurements. In simulation, the model uses its own outputs computed at preceding steps. The same distinction analogously holds for state-space models.

\paragraph{Assumptions and identification objective.}
The framework considered in this paper is subject to the following assumptions:
\begin{itemize}
\item[(A1)] The output $y$ by which a model is scored is a scalar; the state vector may have several components, all of which are identified and simulated.
\item[(A2)] All states are measured or can be constructed directly from measured signals, so that no state estimation is required.
\item[(A3)] The measurement equation (\ref{eq:meas}) is given and is not estimated.
\end{itemize}

The identification task is to determine the expression $f$ in (\ref{eq:static}), $g_c$ in (\ref{eq:ct}), or $g_d$ in (\ref{eq:dt}), including the NARX representation (\ref{eq:narx}). The identified model must be accurate on test data under the specified evaluation mode (Section~\ref{sec:eval}), satisfy the prescribed physical requirements, and be suitable for its intended use.
The identified expressions must be closed-form, consisting of finite combinations of the variables, estimated parameters, fixed constants, and permitted mathematical functions. For dynamic systems, this restriction applies to the state-derivative or state-update equations, whereas computing the resulting trajectories may require numerical integration.

\section{The ASPIRE Framework}
\label{sec:method}

\subsection{Overview}
\label{sec:overview}

\begin{figure*}[pos=t]
\centering
\includegraphics[width=\textwidth]{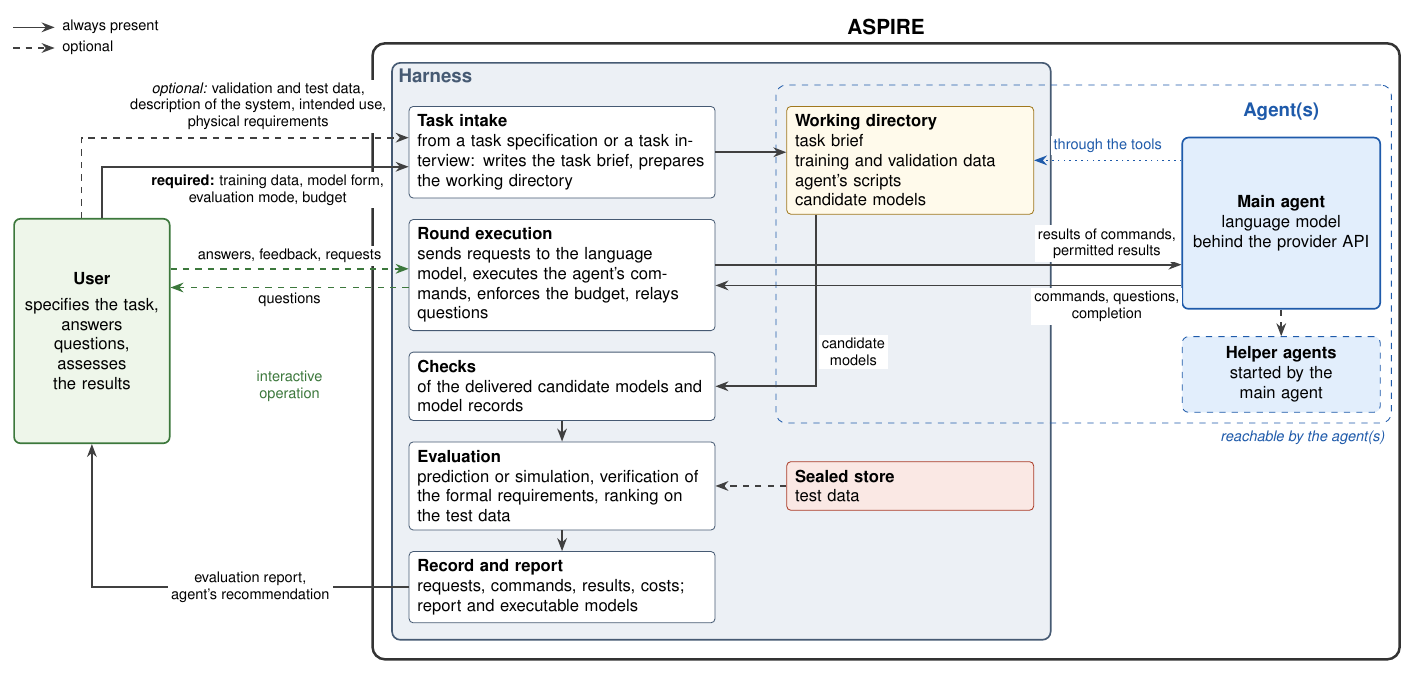}
\caption{Overview of ASPIRE, which consists of the harness (gray) and the agent(s) (blue). Solid arrows and boxes are always present. Dashed ones are optional: the optional part of the task specification, the interactive operation (green) and the helper agents. The agent acts on the working directory only through the tools that the harness executes (dashed blue boundary) and never reaches the sealed test data.}
\label{fig:overview}
\end{figure*}

ASPIRE consists of a \emph{harness} and one or more \emph{agents} (Figure~\ref{fig:overview}). The agent is a language model that carries out the identification: it inspects the data, proposes model structures, writes and runs estimation code, and explains its models. The harness is the software around the agent: it passes the task to the agent, executes the agent's commands in a working directory, enforces a budget, checks and evaluates the delivered models, and stores a complete record of the session. The \emph{user} specifies the task, answers the agent's questions in interactive operation, and assesses the results. The test data stay with the harness and are \emph{sealed}, that is, inaccessible through the agent's files, tools, and conversation.

One attempt by the agent to produce or revise a set of models within the budget is a \emph{round}, and the set of models delivered at the end of that round is the \emph{pool}. A \emph{session} consists of one round in autonomous operation, and of several rounds with analyses between them in interactive operation (Section~\ref{sec:interactive}). After each round, the user receives the models with their errors on the training, validation, and test data, the outcomes of the physical requirements, the agent's recommendation, and the models as executable code. The harness presents these results as a report for reading in a browser (Section~\ref{sec:res-records}).

\subsection{Task specification}
\label{sec:task}

A task has a compulsory and an optional part. The user must provide the training data, the model form of Section~\ref{sec:problem} with the roles of the data columns, the evaluation mode (Section~\ref{sec:eval}) and the budget; a dynamic task also needs the sampling period and the measurement equation (\ref{eq:meas}). Everything else is optional: validation data, test data, a description of the system and of the data, the intended use of the model, physical requirements, known constants, and references to model families from the literature. The division of the data into training, validation, and test data is made by the user. The agent receives this information as a \emph{task brief}, a text that the harness generates from the specification by fixed rules, without a language model, and that also states what the user did not provide.

By default, a model may contain the arithmetic operations, powers, the elementary functions ($\exp$, $\log$, the square root, the trigonometric and hyperbolic functions and their inverses), the absolute value, the sign, the minimum and the maximum, and piecewise definitions with relational and logical conditions; the user may adapt this set. Every value obtained by parameter estimation is a named parameter, whereas a numeric literal that belongs to the form of an expression, such as the exponent in $x^2$, is not.

The user states physical knowledge either as formal requirements or in natural language. A formal requirement has one of the nine types of Table~\ref{tab:requirements}, which consolidates the constraint types of the physics-aware SR methods of~\cite{kubalik2020gecco,kubalik2021eswa,kubalik2023n4sr}, and it specifies the expression concerned, the domain, and the tolerance. For example, the states of the friction system of Section~\ref{sec:task-friction} are an angle $\theta$ and an angular velocity $\omega$, and a requirement of the type equation, $\mathrm{d}\theta/\mathrm{d}t - \omega = 0$, ensures that every delivered model keeps this kinematic relation. The harness verifies the formal requirements on every delivered model (Section~\ref{sec:eval}), and by default, a demonstrated violation excludes the model from the ranking and from the recommendation. Requirements in natural language are passed to the agent verbatim and are assessed by the user. The user either states the requirements in natural language or enters necessary conditions of the nine types that follow from it, each on a stated domain; Table~\ref{tab:friction-conditions} gives an example of formal requirements, and Appendix~\ref{app:requirements} gives an example of requirements in natural language.
\begin{table}[htbp]
\centering
\caption{Types of formal physical requirements.}
\label{tab:requirements}
\begin{tabular}{p{0.27\columnwidth}p{0.62\columnwidth}}
\toprule
Type & Statement \\
\midrule
Symmetry & invariance under a permutation of variables \\
Monotonicity & sign of the partial derivative with respect to one variable on a domain \\
Parity & odd or even under a sign change of one or several variables jointly \\
Exact value & value at a specified point, with a tolerance \\
Inequality & relation between the output and the variables on a domain \\
Curvature & convexity or concavity with respect to one variable \\
Limit & value as a variable tends to a limit \\
Equation & an expression that must equal zero, possibly at a point \\
Relation at~points & a relation required at a specified set of samples \\
\bottomrule
\end{tabular}
\end{table}

\subsection{Candidate models and the model record}
\label{sec:candidates}

The agent is instructed to explore different identification approaches, such as physically motivated modeling, symbolic regression (SR), and sparse regression, and to deliver a \emph{pool} of candidate models with distinct structures rather than variations of the same model form. The maximum number of candidates is specified by the user. The agent may recommend one candidate based on the training and validation data, and this recommendation is recorded before any test results become available.

Each \emph{candidate} consists of the model equations and a \emph{model record} with three parts: a parameter table, an estimation record, and a discussion.

The \emph{parameter table} lists every named quantity in the equations, specifying its value, unit, type, and description. Three parameter types are distinguished: \emph{fitted}, for values estimated from data; \emph{fixed}, for values specified independently of estimation, with their source identified (the task, physics, literature, normalization, or a modeling choice); and \emph{tied}, for values computed from other parameters using a stated equation.

The \emph{estimation record} specifies the estimation criterion, the data used, the estimation method, and its starting values, together with the reasons for these choices. It also identifies the script that reproduces the estimated values.

The \emph{discussion} identifies the model family, interprets each term in the equations, explains how the model structure was obtained, and describes the model's limitations. Terms without a known physical interpretation are characterized by their mathematical effects. The account of model construction distinguishes structures derived from physical reasoning, obtained through search, or proposed as hypotheses, and documents the alternatives considered and the reasons for subsequent modifications.

The model record follows a common format across all tasks. The agent receives a template with an example based on an artificial system, along with instructions to define every symbol, specify units for numerical values, cite only verifiable references, justify modeling choices, and avoid unsupported physical interpretations. Section~\ref{sec:res-records} summarizes a delivered record, and Appendix~\ref{app:records} gives three delivered models with their parameter tables.

The harness checks the completeness and consistency of each record with the model equations. In particular, all symbols must be declared, all parameters must be used, and the expressions must be real and finite on the training data. These checks do not establish whether the reported estimation procedure was actually followed or whether a parameter declared as fitted was estimated as stated. The harness, therefore, retains the agent's scripts and execution logs, allowing the reported identification procedure to be checked against the actual computations.

\subsection{Agent and budget}
\label{sec:execution}

The agent may read and write files in its working directory, write and run code, search the web, install software packages, and delegate subtasks to helper agents, which are further instances of the language model. Its environment contains Python with numerical libraries, among them SciPy~\cite{virtanen2020scipy}, the SR package gplearn~\cite{stephens2026gplearn}, and the sparse-regression package PySINDy~\cite{desilva2020pysindy}; whether it uses them is its decision. Models of any kind, including black-box models, may serve as intermediate tools, but only closed-form models are to be delivered. The instruction that the agent receives with every request -- the system prompt -- is given in Appendix~\ref{app:prompt}. No tool gives access to the test data or to evaluation results. Restricting the agent, for example, to forbid web access, can be stated in the task description.

An identification round is summarized in Algorithm~\ref{alg:round}. A \emph{turn} is one response of the language model, which either requests actions or declares the pool complete. A round is limited in the number of turns, in the execution time, and in the cost of the language model; the default limits are 60 turns, 3600\,s, and 10 US dollars. When the agent declares the pool complete, the harness checks the candidates; if none pass, the agent receives the reasons and may correct the pool once. When a limit is reached, the harness evaluates all candidates delivered so far to ensure valid models are not discarded. The harness records every request, response, command, and result of a session, along with the language model identifier, software versions, and cost.

\begin{algorithm}[htbp]
\linespread{1}\selectfont
\caption{One identification round. The harness executes the steps; the agent produces the responses.}
\label{alg:round}
\begin{algorithmic}[1]
\State give the agent the task brief and the training and validation data
\While{no budget limit is reached}
  \State obtain the next response of the agent \Comment{one turn}
  \If{the response requests actions}
    \State execute them in the working directory and return the results
  \ElsIf{the response declares the pool complete}
    \State check the candidates
    \If{a candidate passes or a correction was already requested} \State \textbf{break} \EndIf
    \State return the reasons for the rejection
  \EndIf
\EndWhile
\State record the recommendation; evaluate every complete candidate; verify the formal requirements
\end{algorithmic}
\end{algorithm}

\subsection{Evaluation}
\label{sec:eval}

The harness computes all predictions used in the evaluation, with the same settings for all candidates. A static model is evaluated algebraically. A dynamic model is evaluated either by one-step-ahead prediction from the measured state or by simulation, as specified in the task. In simulation, the state is initialized once per trajectory, and the model state is propagated using the recorded inputs without correction from later measurements. By default, a continuous-time model is integrated by the fourth-order Runge--Kutta method, with one step per sampling period and the input held constant over the step.

A simulation error grows with the simulated length, so that simulation errors on trajectories of different lengths are not comparable. If the test trajectories are shorter than the training and validation trajectories, the user may therefore specify a horizon $H$, the number of sampling periods simulated in a test trajectory. The harness then divides every training and validation trajectory into consecutive segments of $H$ sampling periods that do not overlap; the last segment of a trajectory may be shorter. Each segment is simulated on its own: the state is initialized from the measured state at the first sample of the segment and is propagated to the end of the segment without correction. The starting points of the simulations are thus $H$ samples apart. This differs from a receding horizon, in which a simulation of $H$ samples would start at every sample. Every sample after the initialization of a trajectory is scored exactly once. The training and validation errors are the root mean squared errors (RMSEs) pooled over all segments, computed using (\ref{eq:score}) with the segments in place of the test trajectories. The test trajectories are not divided.

The candidates are ranked by the RMSE of the output, pooled over the test trajectories $j = 1, \dots, J$,
\begin{equation}
S = \sqrt{\frac{\sum_{j=1}^{J} \sum_{k \in I_j} \left(\hat{y}_j[k] - y_j[k]\right)^2}{\sum_{j=1}^{J} N_j}},
\label{eq:score}
\end{equation}
where $y_j[k]$ is the measured output, $\hat{y}_j[k]$ the predicted or simulated output, $I_j$ the set of scored samples of trajectory $j$ and $N_j$ its size. The test score $S$ has the unit of the output. A candidate is \emph{eligible} for the ranking if it passes the checks of Section~\ref{sec:candidates}, has no demonstrated violation of a formal requirement, and completes the evaluation on every test trajectory with real and finite values.

The \emph{recommendation} of the agent is made without test data. In contrast, the ranking by $S$ uses the test data: it shows whether the pool contains an accurate model, not whether that model can be recognized without test data. We therefore call the lowest $S$ in a pool an \emph{oracle-based selection} and always report it with the pool size.

The harness verifies each formal requirement symbolically where its type permits, and otherwise by sampling its domain. The outcome is a proof, a counterexample or an inconclusive result; a sampled check that finds no violation is inconclusive.

Because language model responses are stochastic, a task is commonly run several times under identical settings, called an \emph{experiment}. For an experiment, we report the number of runs attempted, the number of runs that delivered a pool passing the checks, and the number of runs with at least one eligible candidate. Each quantity is reported as the mean and the sample standard deviation across the runs, along with the cost.

\subsection{Interactive operation}
\label{sec:interactive}

In interactive operation, the user takes an active part throughout the session. During a round, the agent may ask the user a specific question whose answer would affect a modeling choice. If the user cannot answer, the agent states the assumption it makes and how that assumption limits the result. After a round, the user receives the complete evaluation, including the test errors, whereas the agent receives only the results on the training and validation data. The user may then ask the agent for an explanation or a further analysis, or give feedback and request a new round, in which the agent revises the pool; earlier pools are kept. Because the user may have seen test results before giving feedback, every later round is labeled \emph{test-informed} and is not presented as an independent test of the identification.

A session can also start from the user's data files alone, without any accompanying description or metadata. In a \emph{task interview}, an agent examines the training and validation data, proposes what the data show (the roles of the columns, the sampling period, the ranges), asks the user for what the data do not determine (what the system is, what the model will be used for, the units, the physical requirements), and writes the task specification. The user confirms or edits the resulting task brief before the identification starts. The test data are sealed from the interview and from every agent. We use a language model, not a fixed questionnaire, for the interview, so that a term can be explained to the user and an unclear answer can be followed up on.

\section{Experimental Setup}
\label{sec:setup}

We evaluated ASPIRE on three systems with measured data: a mobile robot, a rotational motion under dry friction, and the pressure dynamics of a fermenter (Table~\ref{tab:systems}). All three tasks are dynamic, and the models are evaluated by simulation.\footnote{The harness, the task specifications with their data, and the complete records of all runs are available in a public repository: \repositorylink.}
\begin{table*}[pos=tp]
\centering
\caption{The three test systems. Model form: discrete-time (\ref{eq:dt}) or continuous-time (\ref{eq:ct}). Data: number of trajectories and of samples; for the test data, the number of scored samples, which excludes the first sample of each trajectory.}
\label{tab:systems}
\footnotesize
\setlength{\tabcolsep}{4pt}
\begin{tabular}{L{0.19\textwidth}L{0.24\textwidth}L{0.24\textwidth}L{0.24\textwidth}}
\toprule
 & Mobile robot & Friction system & Fermenter \\
\midrule
Model form & discrete-time & continuous-time & continuous-time \\
States & position $x$, $y$ (m), heading $\phi$ (rad) & angle $\theta$ (rad), angular velocity $\omega$ (rad/s) & pressure $P$ (bar) \\
Inputs & forward velocity $v_f$ (m/s), angular velocity $v_a$ (rad/s) & motor voltage $u$ (V) & valve position $u_\mathrm{v}$ (\% closed), air flow rate $u_\mathrm{f}$ (l/min) \\
Output & $x$ & $\theta$ & $P$ \\
Sampling period $T_s$ & 0.2\,s & 0.01\,s & 5\,s \\
Training data & 3 trajectories, 450 samples & 13 trajectories, 3522 samples & 1 trajectory, 1452 samples \\
Validation data & none & 3 trajectories, 1192 samples & 1 trajectory, 1454 samples \\
Test data & 12 trajectories, 298 scored samples & 4 trajectories, 978 scored samples & 1 trajectory, 1209 scored samples \\
Horizon $H$ & 25 samples & none & none \\
Formal requirements & none & R1--R3 & none \\
Requirements in natural language & none & none & R1--R5 \\
\bottomrule
\end{tabular}
\end{table*}

\subsection{Mobile robot}
\label{sec:task-turtlebot}

The experimental platform is a TurtleBot~2 mobile robot with a differential drive, which moves on a flat floor. The state is the pose of the robot logged by the on-board odometry: the position coordinates $x$ and $y$ in meters and the heading $\phi$ in radians, where $\phi = 0$ is the direction of the $x$ axis and the logged heading is wrapped to the interval $(-\pi, \pi]$. The input consists of the two commands sent to the robot: the forward velocity $v_f$ in m/s and the angular velocity $v_a$ in rad/s. For this system, $x$ and $y$ denote the two position coordinates and not the state vector and the output of Section~\ref{sec:problem}: in the notation of that section, the state vector is $[x, y, \phi]^\top$, the input vector is $[v_f, v_a]^\top$, and the measured output is the position $x$. The data columns have the names \texttt{x}, \texttt{y}, \texttt{phi}, \texttt{v\_f}, and \texttt{v\_a}.
The agent provides the three components of $g_d$ in (\ref{eq:dt}), one expression for the next value of each state. The commands were drawn at random and held for five samples each. The logged pose carries a considerable measurement error of unknown magnitude, and the heading responds to the angular velocity command with a delay of about one sample, which a model of this form cannot represent.

The data are five trajectories of 150 samples each, recorded with a sampling period $T_s = 0.2\;\mathrm{s}$. Three trajectories are the training data. The other two are split into 12 test trajectories of 25 samples (5\,s), in accordance with the intended use stated in the task: simulating the position $x$ over about 5\,s from one logged pose and the sequence of commands. No physical requirements are stated.

\subsection{Rotational motion under dry friction}
\label{sec:task-friction}

The second system is a laboratory setup~\cite{kubalik2026ansr} in which a voltage-controlled direct-current motor drives a vertical disc with an off-center weight, so that gravity acts on the rotation. A leaf spring presses a felt pad against the disc, adding dry friction to the bearings. We refer to it as the friction system. The data contain four signals. The state vector is $[\theta, \omega]^\top$, where $\theta$ is the angle of the disc in radians (column \texttt{theta}), measured by an encoder, with $\theta = 0$ at the stable equilibrium, and $\omega$ is the angular velocity in rad/s (column \texttt{omega}). The input is the motor voltage $u$ in volts (column \texttt{u}), a scalar. The fourth signal is the angular acceleration $\mathrm{d}\omega/\mathrm{d}t$ in rad/s$^2$ (column \texttt{domega}); it serves as a state derivative and is provided as a possible fitting target. The angular velocity and the angular acceleration are obtained from the measured angle by numerical differentiation. The measured output is the angle $\theta$. The agent provides the two components of $g_c$ in (\ref{eq:ct}), the expressions for $\mathrm{d}\theta/\mathrm{d}t$ and $\mathrm{d}\omega/\mathrm{d}t$. The data consist of driven trajectories, which start at rest under a random binary voltage, and of free decays, which start from an initial angle with $u = 0$ and end with the disc sticking at rest. The description given to the agent names the two sources of friction and the two kinds of trajectories. The intended use is to simulate $\theta$ over a full trajectory, with stable behavior over a long horizon and realistic stick-slip behavior.

The task states three physical requirements, all of them formal (Table~\ref{tab:friction-conditions}). We denote by $f(\theta, \omega, u)$ the expression for $\mathrm{d}\omega/\mathrm{d}t$. R1 is the kinematic relation $\mathrm{d}\theta/\mathrm{d}t = \omega$. R2 states that gravity acts toward the stable equilibrium, whereas static friction may prevent motion: at rest and without voltage, the angular acceleration is zero or points toward $\theta = 0$, that is, $f(\theta, 0, 0)\,\theta \leq 0$ for every angle. R3 states the influence of the control input: $f$ does not decrease as the voltage increases; it may remain constant, as it does when the system sticks due to friction. The tolerance of every requirement is zero, and a demonstrated violation excludes a model. The brief gives the agent the three conditions with their domains, as in Table~\ref{tab:friction-conditions}, without a verbal statement.

\begin{table*}[pos=tp]
\centering
\caption{Formal requirements of the described task of the friction system. $f(\theta, \omega, u)$ denotes the delivered expression for $\mathrm{d}\omega/\mathrm{d}t$ in rad/s$^2$, with $\theta$ in rad, $\omega$ in rad/s and $u$ in V. Types as in Table~\ref{tab:requirements}. The tolerance of every requirement is zero.}
\label{tab:friction-conditions}
\footnotesize
\setlength{\tabcolsep}{4pt}
\begin{tabular}{lllL{0.45\textwidth}}
\toprule
Requirement & Type & Condition & Domain \\
\midrule
R1 & equation & $\mathrm{d}\theta/\mathrm{d}t = \omega$ & $\theta \in [-3.1416, 3.1416]$, $\omega \in [-23, 23]$, $u \in [-10, 10]$ \\
R2 & inequality & $f\,\theta \leq 0$ & $\theta \in [-\pi, \pi]$, $\omega = 0$, $u = 0$ \\
R3 & monotonicity & $f$ non-decreasing in $u$ & $\theta \in [-3.1416, 3.1416]$, $\omega \in [-23, 23]$, $u \in [-10, 10]$ \\
\bottomrule
\end{tabular}
\end{table*}

\subsection{Pressure dynamics of a fermenter}
\label{sec:task-fermenter}

A 40-liter laboratory fermenter is filled with about 25 liters of water~\cite{vancan1995design,babuvska2012fuzzy,kubalik2020gecco}. Air is fed from the bottom at a flow rate set by a mass-flow controller and exits through an outlet valve at the top, so the pressure in the headspace depends on the airflow rate and the valve position. The data contain four signals. The only state is the air pressure in the headspace in bar (column \texttt{p}), which is also the measured output; we denote it by $P$ to keep $p$ for the parameter vector. The input consists of two signals: the position of the outlet valve $u_\mathrm{v}$ in percent closed (column \texttt{valve}), where 100\,\% is fully closed, and the air flow rate $u_\mathrm{f}$ in l/min (column \texttt{flow}). The fourth signal is the time derivative $\mathrm{d}P/\mathrm{d}t$ in bar/s (column \texttt{dp}), computed from the pressure using a filtered numerical difference; it serves as a state derivative and is provided as a possible fitting target. The pressure is quantized. The agent provides $g_c$ in (\ref{eq:ct}), which here is the single expression for $\mathrm{d}P/\mathrm{d}t$. The training, validation, and test data are each a single trajectory, so the test score is based on a single trajectory. 
The intended use is the simulation of the pressure over a whole trajectory, with stable behavior over a long horizon and with a model whose structure can be defended from the physics and not only from the fit.

The task states five physical requirements in natural language: at a fixed pressure, the rate of the pressure does not decrease when the air flow rate increases (R1) or when the valve is closed further (R2); with zero air flow and an open valve, atmospheric pressure is the equilibrium (R3) and the pressure tends toward it (R4); and with a fully closed valve the pressure cannot decrease while air is supplied (R5). See Appendix~\ref{app:requirements} for full detail. The agent receives the requirements verbatim. The harness does not verify them; we assessed them on the recommended models as described in Section~\ref{sec:assessment}. No formal requirements are stated.

\subsection{Described and blind tasks}
\label{sec:design}

Every system is run as two tasks. The \emph{described} task contains the information given above. The \emph{blind} task contains the same data (the same dataset split), sampling period, and column roles, and the same evaluation settings, but nothing that tells what the system is: the columns are named $x_1, x_2, \dots$ and $y$, and the units, the description, the intended use, the ranges, and the requirements are omitted.

This allows us to assess the effect of prior knowledge on the identified models. The blind tasks also reduce the possibility that the agent reproduces a known physical model from its training data rather than inferring the model structure from the measurements. Finally, they represent system identification with limited prior knowledge, a common setting in symbolic regression.

\subsection{Compared methods}
\label{sec:references}

We compare the models of the agent with those of two established tools for closed-form models, applied to the training data of each system:
\begin{itemize}
\item PySINDy~\cite{desilva2020pysindy} -- sparse identification of nonlinear dynamics~\cite{brunton2016sindy} with a polynomial library of degree 2 in the states and inputs and sequentially thresholded least squares;
\item PySR~\cite{cranmer2023pysr} -- SR by regularized evolution of expression trees.
\end{itemize}
Each tool is used at two settings. The \emph{default} setting is that of the package: the threshold 0.1 for PySINDy, and for PySR, the four arithmetic operators and the equation that the package selects. The \emph{matched} setting brings the tool closer to the agent's conditions. For PySINDy, the threshold is selected from a grid. For PySR, the unary functions $\sin$, $\cos$, $\tanh$, $\exp$, $\log$, the square root, the absolute value, and the sign are added, and the equation is selected from the Pareto front of accuracy and complexity. Both selections minimize the simulation error on the training data. The tools fit the next value of each state for the mobile robot and the supplied derivative for the other two systems. The harness evaluates their models exactly as the agent's candidates, including verification of the formal requirements. The tools do not receive the requirements or system description. We therefore report the test score for each model and, separately, the number of its models with a demonstrated violation, which would exclude a candidate of the agent. PySINDy is deterministic; PySR is run with five seeds, and the results are reported as the mean and sample standard deviation.

\subsection{Assessment of the model records}
\label{sec:assessment}

The harness checks the form of a model record; it does not check its content (Section~\ref{sec:candidates}). We therefore assessed the record of the recommended candidate of every run, 30 records in total, by five items, each marked \emph{yes}, \emph{partly} or \emph{no}:
\begin{itemize}
\item[(a)] every term of the expressions is interpreted, and the interpretation is correct; for a blind task, a term is described by its mathematical effect, and no physical meaning is invented;
\item[(b)] the stated origin of the structure agrees with the commands and results in the record of the run;
\item[(c)] the declared type of every parameter (fitted, fixed, or tied) agrees with the estimation script;
\item[(d)] the criterion, the estimation method, and the starting values each have a stated reason;
\item[(e)] the stated limitations follow from the data or from the structure of the model.
\end{itemize}
An item is marked \emph{yes} when everything that it concerns satisfies it, \emph{partly} when one term, claim, parameter, reason, or limitation does not, and \emph{no} when several do not. The assessment was performed initially by Claude Code using Opus~5.5 on high effort on 8 October 2026, and manually verified by the authors.

The requirements in natural language for the fermenter were assessed using the recommended model for each of the five runs of the described task, with two marks per requirement. The first mark states whether the discussion in the model record addresses the requirement (\emph{yes} or \emph{no}). The second states whether the delivered expression with the delivered parameter values satisfies the requirement (\emph{yes}, \emph{partly}, \emph{no}, or \emph{not decidable} from the expression). It rests on evaluating the expression on a grid of valve positions from 0 to 100\,\%, air flow rates from 0 to 31\,l/min, and pressures from 0.9 to 2.3\,bar, ranges that contain the training data. R1, R2, and R5 are evaluated at pressures of at least 1\, bar, because below atmospheric pressure, air enters through the outlet valve and the statements do not apply. For R3 and R4, a valve position of at most 90\,\% counts as open, and an equilibrium pressure within 0.05\,bar of 1\,bar counts as atmospheric. A requirement is marked \emph{partly} when it holds on a part of the grid: for R1, R2, and R5 on at least 95\,\% of the grid points, for R3 and R4 at some of the open valve positions.

\subsection{Settings}
\label{sec:settings}

All runs used the language model \texttt{gpt-6.1-sol} of OpenAI~\cite{openai2026gpt61sol} with medium reasoning effort in the role of the agent. The provider has not published a dated snapshot of this model; the runs are dated from 7 to 9 October 2026. Each of the six tasks was run five times, which gives six experiments and 30 runs. Every run was one round in autonomous operation with the default budget of Section~\ref{sec:execution} and a pool of at most 10 candidates. The cost is computed from the provider's prices at the time of the runs: 2 and 10 US dollars per million input and output tokens, respectively.

\section{Results}
\label{sec:results}

All 30 runs ended when the agent declared its pool complete, and no run reached a budget limit; the largest consumption of a run was 52 turns, 2242\,s and 3.38 US dollars. Every run delivered a pool that passed the checks, and all 126 candidates of the 30 pools are eligible: no candidate was excluded by a formal requirement, and no simulation failed. In every experiment, five runs were therefore attempted, five delivered a pool, and five had an eligible candidate. The agent recommended a candidate in every run. A summary of the six experiments can be found in Table~\ref{tab:aspire}.

After every run, the harness scanned the agent's files and conversation for values of the sealed test data. It reported 16 findings across 7 runs, and we examined each. All are numbers that the agent computed from the training or validation data, or copied from them into its own files, and that coincide with a test value in at least six significant digits. No finding indicates access breach to the test data.

\begin{table*}[pos=tp]
\centering
\caption{Results of ASPIRE per experiment: mean $\pm$ sample standard deviation over five runs (five finite values in every entry). Training error: RMSE of the simulated output of the recommended model on the training data, for the mobile robot in segments of $H = 25$ samples. $S$: test score (\ref{eq:score}). Recommended: the model that the agent recommended without test data. Lowest in pool: oracle-based selection on the test data, with the pool sizes of the five runs. Errors in m (mobile robot), rad (friction system) and bar (fermenter). Language model \texttt{gpt-6.1-sol}, runs of 7 to 9 October 2026.}
\label{tab:aspire}
\footnotesize
\setlength{\tabcolsep}{3pt}
\begin{tabular}{llcccccc}
\toprule
System & Task & Training error, & $S$, & $S$, lowest & Pool sizes & Fitted param., & Cost per run \\
 & & recommended & recommended & in pool & & recommended & (US dollars) \\
\midrule
Mobile robot & described & 0.0397 $\pm$ 0.0007 & 0.0430 $\pm$ 0.0010 & 0.0382 $\pm$ 0.0016 & 6, 4, 6, 5, 4 & 2.4 $\pm$ 0.9 & 1.31 $\pm$ 0.25 \\
Mobile robot & blind & 0.0394 $\pm$ 0.0014 & 0.0401 $\pm$ 0.0015 & 0.0365 $\pm$ 0.0019 & 3, 3, 4, 4, 6 & 9.6 $\pm$ 5.9 & 1.91 $\pm$ 0.47 \\
Friction system & described & 0.0511 $\pm$ 0.0085 & 0.0417 $\pm$ 0.0070 & 0.0279 $\pm$ 0.0075 & 6, 3, 6, 4, 3 & 7.6 $\pm$ 1.7 & 2.20 $\pm$ 0.50 \\
Friction system & blind & 0.0459 $\pm$ 0.0096 & 0.0408 $\pm$ 0.0079 & 0.0387 $\pm$ 0.0067 & 3, 3, 3, 4, 3 & 8.4 $\pm$ 1.5 & 2.09 $\pm$ 0.45 \\
Fermenter & described & 0.0253 $\pm$ 0.0008 & 0.0232 $\pm$ 0.0006 & 0.0231 $\pm$ 0.0005 & 4, 3, 4, 5, 3 & 5.6 $\pm$ 1.1 & 2.01 $\pm$ 0.45 \\
Fermenter & blind & 0.0250 $\pm$ 0.0024 & 0.0225 $\pm$ 0.0010 & 0.0223 $\pm$ 0.0007 & 4, 6, 5, 5, 4 & 19.0 $\pm$ 2.2 & 2.70 $\pm$ 0.44 \\
\bottomrule
\end{tabular}
\end{table*}

\subsection{Comparison with PySINDy and PySR}
\label{sec:res-accuracy}

Table~\ref{tab:methods} shows the test scores of the models recommended by ASPIRE with those of the models found by PySINDy and PySR.

\begin{table*}[pos=tp]
\centering
\caption{Test score $S$ (\ref{eq:score}) and number of fitted parameters of the models recommended by the agent and of the models found by PySINDy and PySR (Section~\ref{sec:references}), evaluated by the harness on the same test data. ASPIRE: mean $\pm$ sample standard deviation over five runs; PySR: over five seeds, with the range of the number of parameters; PySINDy: a single deterministic model. The lowest mean $S$ of each system is set in bold; no statistical test was applied. $^\dagger$: every model of the tool violates the formal requirement R2 of the described task of the friction system (Table~\ref{tab:friction-conditions}); a candidate of the agent with such a violation is excluded.}
\label{tab:methods}
\footnotesize
\setlength{\tabcolsep}{3pt}
\begin{tabular}{lcccccc}
\toprule
 & \multicolumn{2}{c}{Mobile robot} & \multicolumn{2}{c}{Friction system} & \multicolumn{2}{c}{Fermenter} \\
Method & Param. & $S$ (m) & Param. & $S$ (rad) & Param. & $S$ (bar) \\
\midrule
ASPIRE, described task & 2.4 $\pm$ 0.9 & 0.0430 $\pm$ 0.0010 & 7.6 $\pm$ 1.7 & 0.0417 $\pm$ 0.0070 & 5.6 $\pm$ 1.1 & 0.0232 $\pm$ 0.0006 \\
ASPIRE, blind task & 9.6 $\pm$ 5.9 & \textbf{0.0401} $\pm$ 0.0015 & 8.4 $\pm$ 1.5 & 0.0408 $\pm$ 0.0079 & 19.0 $\pm$ 2.2 & \textbf{0.0225} $\pm$ 0.0010 \\
PySINDy, default & 3 & 0.2140 & 8 & 0.2668$^\dagger$ & 0 & 0.2631 \\
PySINDy, matched & 55 & 0.1530 & 10 & 0.2692$^\dagger$ & 10 & 0.0284 \\
PySR, default & 6--8 & 0.0504 $\pm$ 0.0068 & 4 & 0.1440 $\pm$ 0.0010$^\dagger$ & 3--5 & 0.0418 $\pm$ 0.0054 \\
PySR, matched & 8--14 & 0.0558 $\pm$ 0.0129 & 5--6 & \textbf{0.0363} $\pm$ 0.0039$^\dagger$ & 1--5 & 0.0748 $\pm$ 0.0632 \\
\bottomrule
\end{tabular}
\end{table*}

\emph{Mobile robot.} The test scores of the ten recommended models lie between 0.0379 and 0.0443\,m, below those of all five PySR models at the default setting (0.0458 to 0.0612\,m). In the matched setting, PySR has a lower training error than in the default setting (0.0263 vs. 0.0654\,m) and a higher test score; one of its five models (0.0438\,m) lies within the range of the agent's models. This indicates that selection based on simulation error across three training trajectories favors equations that do not generalize to the test data. PySINDy at the default threshold removes every term that contains a command, so that its model predicts almost no motion (0.2140\,m); with the selected threshold, it keeps 55 polynomial terms (0.1530\,m).

\emph{Friction system.} The test scores of the five recommended models of the described task lie between 0.0292 and 0.0459\,rad, and those of the blind task between 0.0322 and 0.0492\, rad. PySR at the matched setting (Section~\ref{sec:references}), in which the unary functions are added to the arithmetic operators, finds a sine of the angle and a hyperbolic tangent of the angular velocity and reaches $0.0363 \pm 0.0039$\,rad (0.0335 to 0.0431\,rad over the five seeds). Its mean lies 0.0054\,rad below that of the described task and 0.0045\,rad below that of the blind task. Both differences are smaller than the sample standard deviation of the agent's runs (0.0070 and 0.0079\,rad), and we observe no difference between the three. The models differ in another respect. All 12 models of the two tools violate the requirement R2 of the described task: each contains a constant term or a polynomial in the angle, so that at rest and without voltage, its acceleration points away from the equilibrium at some angles. For the PySR models in the matched setting, the violation is small: for the counterexamples the harness found, at angles of 0.05 to 0.10\,rad, the product $f\theta$ lies between 0.008 and 0.071\,rad$^2$/s$^2$. Each of the 12 models would nevertheless be excluded as a candidate of the agent, whereas none of the 22 candidates of the agent in the described task has a demonstrated violation (Section~\ref{sec:res-approaches}). The other three models cannot express the gravity torque: PySR at the default setting has only the four arithmetic operators, and PySINDy uses a polynomial library of degree 2 at both settings, because its matched setting selects only the threshold. These models reach 0.144 to 0.269\,rad.

\emph{Fermenter.} The largest test score of the ten recommended models is 0.0239\,bar, below the score of every model of the two tools. The best of these is the PySINDy model with the selected threshold, which keeps all ten polynomial terms (0.0284\,bar). PySR at the matched setting varies between 0.0276 and 0.1850\,bar across the five seeds: equations with a small error in the derivative drift or leave the data range in the simulation.

\begin{figure*}[pos=tp]
\centering
\includegraphics[width=\textwidth]{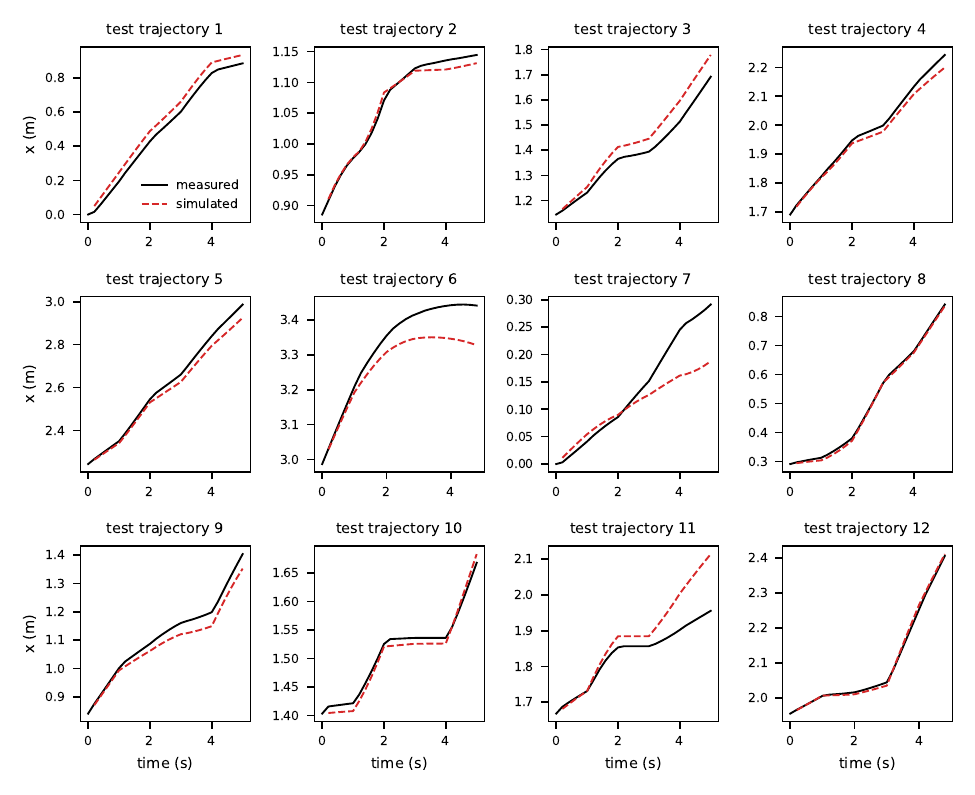}
\caption{Mobile robot, described task: measured (solid) and simulated (dashed) position $x$ on the 12 test trajectories for the recommended model of the run with the median test score ($S = 0.0428$\,m). Each simulation starts from the measured pose and uses the recorded commands without correction.}
\label{fig:test-robot}
\end{figure*}

\begin{figure*}[pos=tp]
\centering
\includegraphics[width=\textwidth]{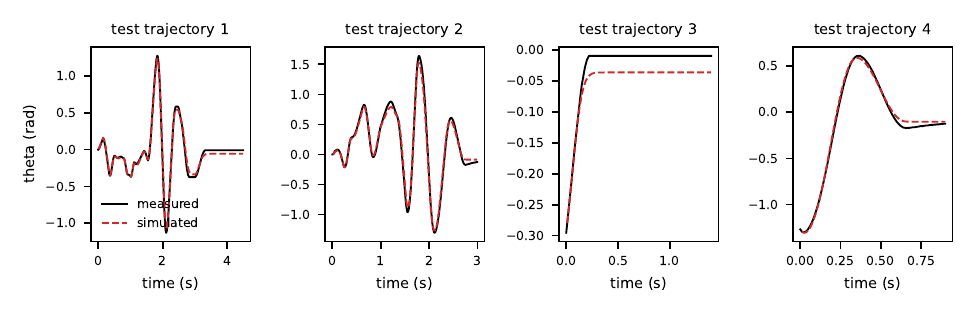}\\[1ex]
\includegraphics[width=\textwidth]{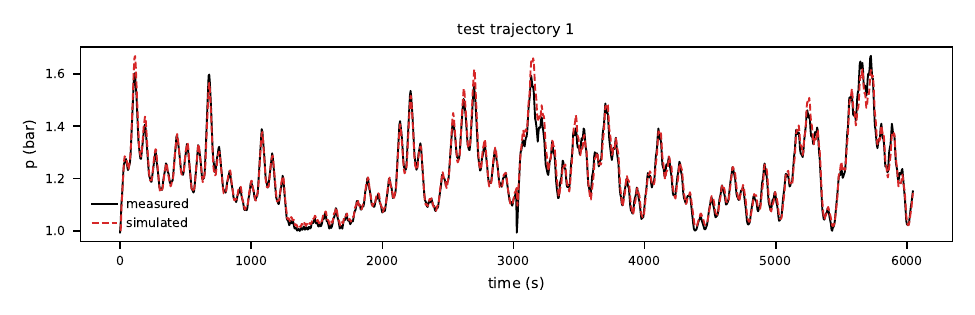}
\caption{Measured (solid) and simulated (dashed) output on the test trajectories for the recommended model of the run with the median test score, described tasks. Top: friction system, angle $\theta$ ($S = 0.0442$\,rad); trajectories 1 and 2 are driven, 3 and 4 are free decays. Bottom: fermenter, pressure $P$ ($S = 0.0229$\,bar).}
\label{fig:test-continuous}
\end{figure*}

Figures~\ref{fig:test-robot} and~\ref{fig:test-continuous} show the measured and simulated outputs on the test data for one recommended model per described task. The run shown is the one whose recommended model has the median test score of the five runs; we fixed this rule before inspecting the results.

\subsection{Delivered models}
\label{sec:res-approaches}

Among the resulting pools in all experiments, each pool contains 3 to 6 candidates, and within each pool, each candidate has its own structure.

Formal requirements are stated only in the described task of the friction system. The harness verified the three requirements in Table~\ref{tab:friction-conditions} for each of the 22 candidates across the five runs. No verification demonstrated a violation, so no candidate was excluded. R1 was proved for all 22 candidates. For R2 and R3, the symbolic stage reached a proof for 5 and for 3 candidates. For the other candidates, whose expressions contain minima and maxima, the evaluation at 10000 sampled points found no violation, which the harness reports as unresolved and not as satisfied. In total, 30 of the 66 verifications ended with a proof and 36 without a decision. In all five runs, the agent checked R2 and R3 itself before it declared the pool complete, by evaluating the fitted expressions at 1000 to 100000 sampled points of the stated domains; no check led to a change of a candidate. Four of the five recommended model records also derive the two requirements from the expression (Appendix~\ref{app:records}).

In the blind task of this system, no requirement is stated; all 16 candidates nevertheless give the rate of the output as the second state. To see what the requirements exclude, we applied the requirements of the described task to the five recommended models of the blind task, after renaming their variables. All five satisfy R1 and R3 and violate R2: each contains a constant term or a shift of the angle, so that at rest and without input, its acceleration is not zero at the equilibrium and points away from it on one side.

No run executed gplearn or PySINDy, the two packages for the search of closed-form models that were installed in the agent's environment (Section~\ref{sec:execution}) and named in the system prompt (Appendix~\ref{app:prompt}). The agent fitted structures that it wrote itself, in several runs by least-squares regression on a library of candidate terms of its own choice. It started a helper agent in 27 of the 30 runs, searched the web in 27 runs and installed a package, a compiler for numerical Python code, in 14 runs.

With the description, the recommended models are physical structures in all 15 runs. For the mobile robot, they are the kinematic model of a unicycle with one gain per command. For the friction system, they combine a gravity torque proportional to the sine of the angle, in three runs with a second harmonic, viscous friction, Coulomb friction, and the voltage drive with a mechanism for sticking at rest. No requirement demands such a mechanism; the description and the intended use name it. In three runs, it is a term that cancels the applied torque at rest up to the friction level and whose weight, a smooth function of the angular velocity, decays over 0.12 to 0.16\,rad/s (Appendix~\ref{app:records}). In the other two runs, the friction term is the applied torque plus a term proportional to the angular velocity, limited to the friction levels, which drives the velocity to zero while the applied torque lies within these levels. Two of the five models have a friction level of its own for each direction of motion, two a friction level that depends on the angle, and one a saturating characteristic of the motor. All five expressions are continuous in the angular velocity. For the fermenter, they are balances of the gas in the headspace with an inflow proportional to the air flow rate, an outflow through a nonlinear resistance and a monotone empirical valve characteristic. Without the description, the agent arrived at the same kind of structure for the mobile robot in three and for the friction system in four of five runs; for the friction system, these models contain, in the neutral names, a sine of the output, a linear term and a smoothed sign of the second state, and a gain of the input. In the other blind runs of these two systems, the recommended models are regression models on trigonometric and polynomial terms (mobile robot, two runs) or replace the sine by a polynomial of degree five (friction system, one run). For the fermenter, all five blind models are third-degree polynomials in the three variables.

\subsection{Model records}
\label{sec:res-records}

\begin{figure*}[pos=tp]
\centering
\includegraphics[width=0.9\textwidth]{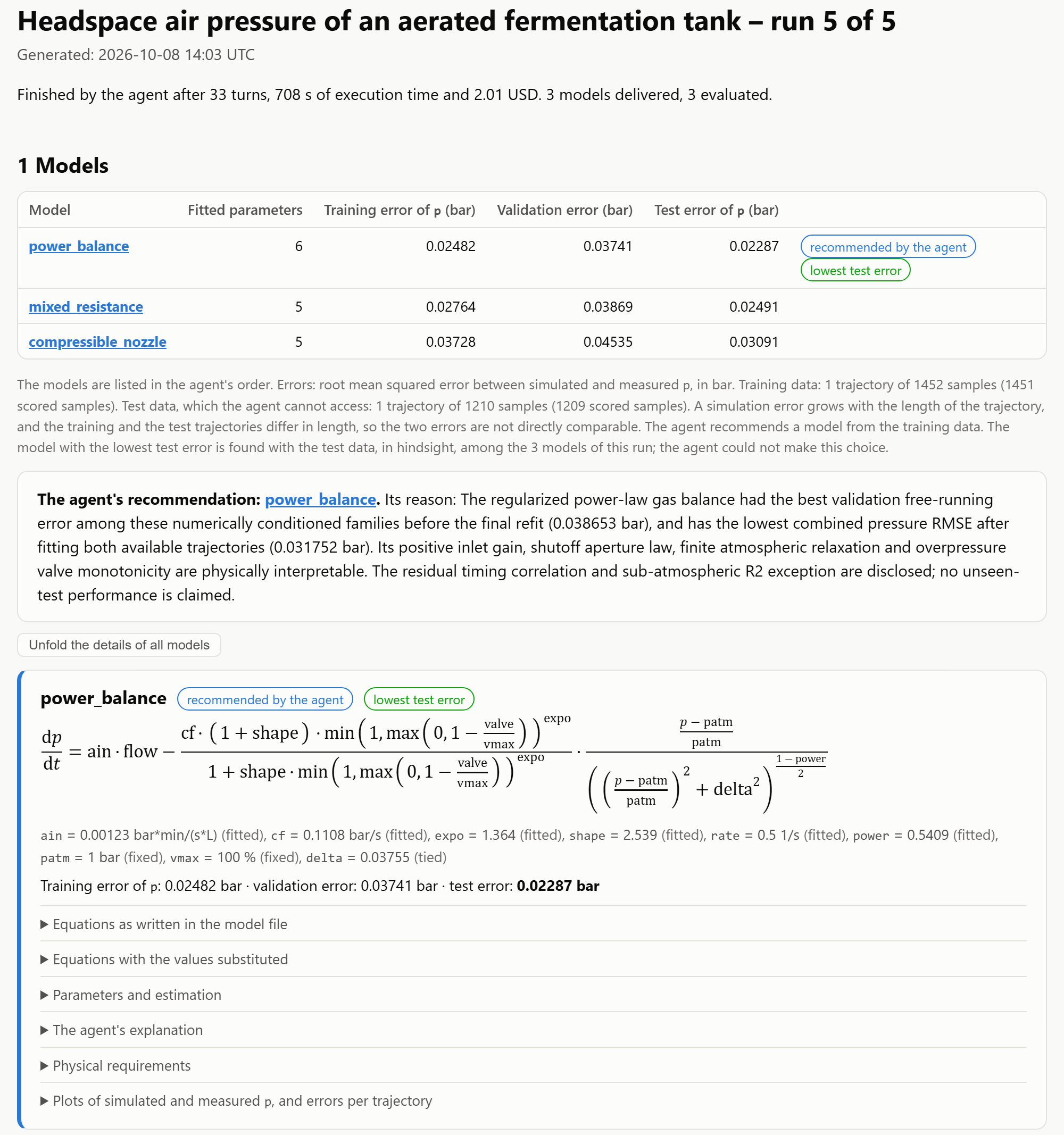}
\caption{Report of a run as shown in the browser: the upper part of the page of the run with the median test score of the described task of the fermenter. Top: the candidates of the pool with the number of fitted parameters and the training, validation, and test errors; the labels mark the candidate recommended by the agent and the candidate with the lowest test error. Middle: the recommendation of the agent with its reason. Bottom: the recommended candidate as a typeset equation with its parameter values and their types. The six lines below it unfold to the equations as written by the agent, the parameter table with the estimation record, the discussion, the physical requirements, and the plots. The names of the variables and of the parameters are those of the agent.}
\label{fig:report}
\end{figure*}

ASPIRE presents the results of every experiment and of every interactive session as a report, a set of HTML pages that are read in a browser. The page of an experiment shows the task brief as the agent received it, the settings, and the statistics over the runs with the recommended model of each run. The page of a run (Figure~\ref{fig:report}) lists the candidates of the pool with their training, validation, and test errors, states the recommendation of the agent with its reason, and shows every candidate as a typeset equation with its parameter values. For each candidate, the reader can unfold the model record, the physical requirements, and the plots of the simulated and the measured output. The page also lists the course of the run turn by turn. Each turn unfolds to the commands of the agent, the files that it wrote, and the outputs that it received, so that a statement of a model record can be compared with what the agent did. The reports of all 30 runs are in the repository and contain the full wording of every model record. We summarize one record here and give the recommended models of the three described tasks in Appendix~\ref{app:records}.

The example is the record of the recommended candidate of the described task of the fermenter in the run with the median test score ($S = 0.0229$\,bar; Figure~\ref{fig:report}), whose model is equation (\ref{eq:rec-fermenter}) with the parameters of Table~\ref{tab:record} in Appendix~\ref{app:records}. The model is a balance of the gas in the headspace with six fitted parameters, two fixed parameters, and one tied parameter. The discussion separates the gas balance, which it takes from physics, from the power law of the outflow and the valve characteristic, which it calls modeling choices. It also states that the textbook it names was cited from memory without checking and that no SR package was run. The estimation record reports three stages, each started from the result of the previous one: the error of the supplied derivative on the training data, the simulation error on the training data, and the simulation error on the training and validation data together. It gives the starting values and the bounds as numbers and adds that the final validation error ``is not an independent generalization estimate''. The limitations address each of the requirements R1 to R5. One parameter shows what the check of the form cannot detect. The relaxation rate $\rho$ is declared as fitted, but its delivered value of 0.5\,s$^{-1}$ equals the upper bound that the agent set so that the numerical integration with the sampling period of 5\,s remains stable (Appendix~\ref{app:records}). The value is therefore determined by a numerical choice of the agent and not by the data alone, which the limitations of the record state.

Table~\ref{tab:assessment} gives the marks of the five items of Section~\ref{sec:assessment} for the 30 recommended candidates. Of the 150 marks, 132 are \emph{yes}, 18 \emph{partly} and none \emph{no}; 13 of the 18 concern records of blind tasks.

\begin{table*}[pos=tp]
\centering
\caption{Assessment of the model records of the recommended candidates: number of records (of five per experiment) marked yes / partly / no for each item of Section~\ref{sec:assessment}: (a) interpretation of every term, (b) agreement of the stated origin with the run record, (c) agreement of the declared parameter types with the estimation script, (d) stated reasons for the criterion, the method and the starting values, (e) limitations that follow from the data or the structure.}
\label{tab:assessment}
\footnotesize
\setlength{\tabcolsep}{4pt}
\begin{tabular}{llccccc}
\toprule
System & Task & (a) & (b) & (c) & (d) & (e) \\
\midrule
Mobile robot & described & 5 / 0 / 0 & 5 / 0 / 0 & 5 / 0 / 0 & 5 / 0 / 0 & 5 / 0 / 0 \\
Mobile robot & blind & 5 / 0 / 0 & 3 / 2 / 0 & 5 / 0 / 0 & 4 / 1 / 0 & 5 / 0 / 0 \\
Friction system & described & 4 / 1 / 0 & 4 / 1 / 0 & 5 / 0 / 0 & 5 / 0 / 0 & 5 / 0 / 0 \\
Friction system & blind & 5 / 0 / 0 & 3 / 2 / 0 & 4 / 1 / 0 & 3 / 2 / 0 & 5 / 0 / 0 \\
Fermenter & described & 5 / 0 / 0 & 2 / 3 / 0 & 5 / 0 / 0 & 5 / 0 / 0 & 5 / 0 / 0 \\
Fermenter & blind & 3 / 2 / 0 & 4 / 1 / 0 & 5 / 0 / 0 & 3 / 2 / 0 & 5 / 0 / 0 \\
\midrule
All & & 27 / 3 / 0 & 21 / 9 / 0 & 29 / 1 / 0 & 25 / 5 / 0 & 30 / 0 / 0 \\
\bottomrule
\end{tabular}
\end{table*}

Half of the marks \emph{partly} concern item (b): in nine records, the stated origin differs from the record of the run in one claim, for example in the order of two steps, or in a comparison that the run made on a part of the data and the record states without this restriction. Five records give no reason for the starting values of the final estimation stage (item (d)), three do not interpret every term (item (a)), and one declares as fitted a parameter that the script holds at a value chosen from a grid of three values (item (c)). In all 30 records, the delivered parameter values equal the values that the script computes. In the blind tasks, no recommended record states a physical meaning of a term as a fact. The evidence for every mark is in the repository.

In 18 of the 20 runs of the friction system and the fermenter, the agent estimated the recommended model on the training and the validation data together, after the validation data had served to compare structures. The records state this, and the training errors of Table~\ref{tab:aspire} are computed by the harness on the training data alone.

For the fermenter, all five recommended models of the described task address each of the requirements R1 to R5 in their record and satisfy each of them on the whole grid of Section~\ref{sec:assessment}. Three records state that R2 does not hold below the equilibrium pressure, where closing the valve reduces the inflow from the atmosphere; the assessment evaluates R2 at pressures of at least 1\,bar.

\subsection{Cost}
\label{sec:res-cost}

A run cost between 0.96 and 3.38 US dollars, the lowest and the highest cost of the 30 runs. Averaged over the five runs of an experiment, a run took 29 to 45 turns and 9 to 21 minutes of execution time, depending on the experiment, and the mean cost of an experiment lies between $1.31 \pm 0.25$ and $2.70 \pm 0.44$ US dollars (Table~\ref{tab:aspire}). The 30 runs cost 61.10 US dollars, which is 0.48 US dollars per eligible candidate. The recorded cost charges cached input tokens at the full price and is therefore an upper estimate. For comparison, one search of PySR took 4 to 31 minutes on one processor core for one fitting target.

\subsection{Interactive case study}
\label{sec:res-interactive}

This section demonstrates the interaction mode capabilities of ASPIRE on an interactive session. The ASPIRE framework offers a terminal and a browser interface. This session was conducted in the browser interface of the harness on 9 October 2026 and started from the data files of the mobile robot alone. The task interview took 15 turns and 0.46 US dollars; the session, with two rounds and one analysis between them, took 57 turns, 17 minutes of execution time and 5.52 US dollars. The interaction provided the following.

\begin{figure*}[pos=tp]
\centering
\includegraphics[width=\textwidth]{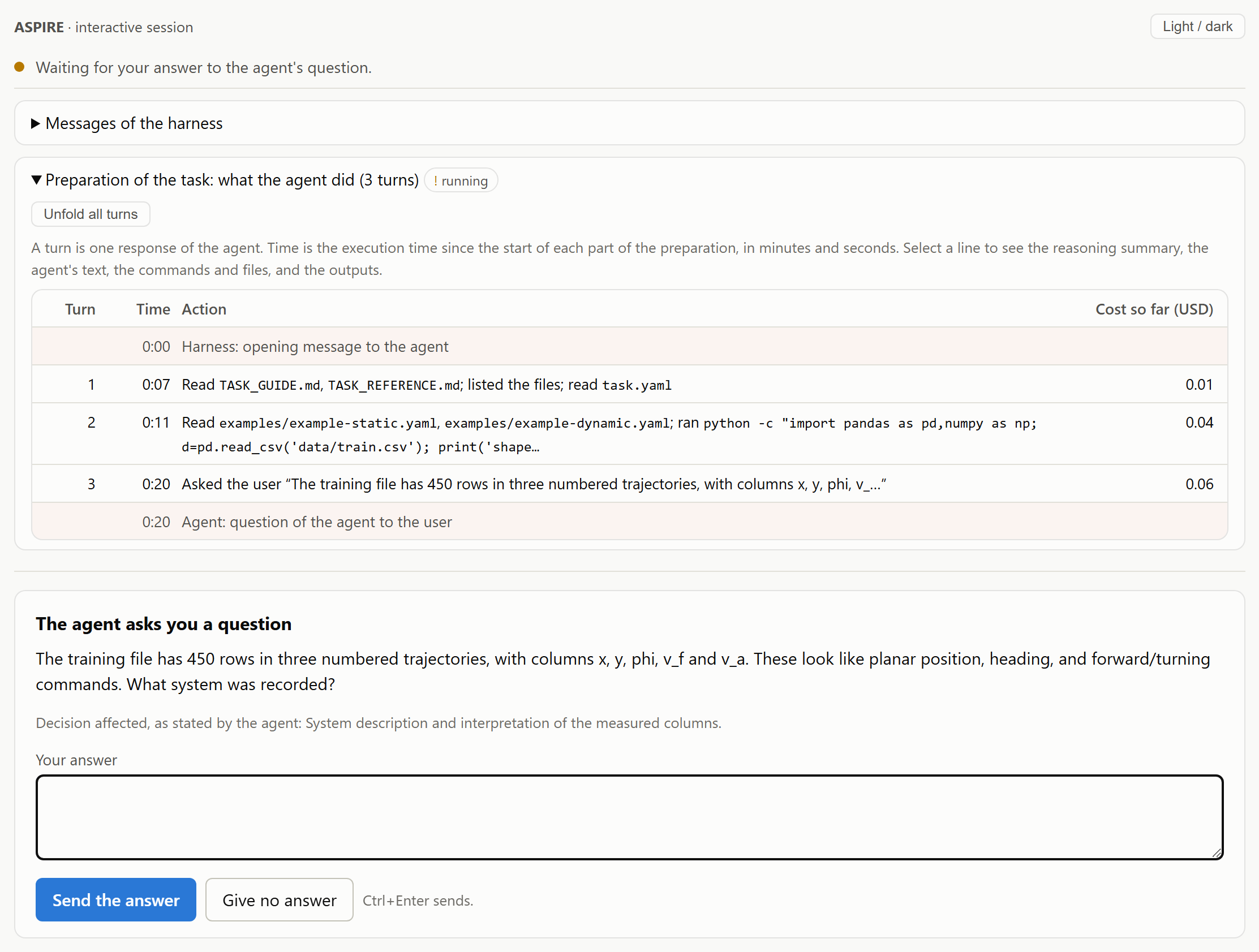}
\caption{ASPIRE's browser interface during the task interview of the interactive session. Top: the status line, which states what the session waits for. Middle: the course of the interview so far, one line per turn of the interview agent with the execution time and the accumulated cost; in three turns, the agent read the guide to the task specification, two example specifications and the training data, and asked its first question. Bottom: the question, the decision that it affects as stated by the agent, and the form in which the user answers or declines to answer.}
\label{fig:interview}
\end{figure*}

\begin{enumerate}
\item \emph{Task specification from data files.} From the training data, the interview agent recognized the trajectories and the probable meaning of the columns, asked six questions about what the data do not determine (Figure~\ref{fig:interview}), and proposed the model form and the evaluation. The resulting specification agrees with the one written by hand (Section~\ref{sec:task-turtlebot}) in the model form, the scored output and the evaluation; the user needs no knowledge of its format.
\item \emph{Questions of the agent.} In round~1, the agent asked about the nature of the measurement error and about the initial states of the intended simulations, each with the modeling decision that it affects. The first answer added that transient effects follow every change of the commands, which the task description did not state; every model record of the round names this among its limitations.
\item \emph{Analysis without a change of the pool.} Asked how much of the error arises after a change of the commands, the agent computed that the first sample after a change, one fifth of the samples, carries 83\,\% of the squared one-step prediction error of the heading. The user can thus test a conjecture on the data, here the delay stated in Section~\ref{sec:task-turtlebot}, before deciding how to proceed.
\item \emph{Feedback and a further round.} The user asked for expressions without a case distinction and for the evidence for two terms, a cubic term and a dead zone of the angular velocity command. The agent reported that 5 of the 450 training samples lie inside the dead zone, withdrew both models, and delivered a model with two fitted parameters as the recorded successor of the dead-zone model; the pool changed from four models to three. What the user considers readable and defensible thereby shapes the pool.
\item \emph{Changes of the task.} With the feedback, the user added two requirements in natural language, which the two revised model records address; a formal requirement can likewise be relaxed or restored. Requirements that become apparent only when models are seen need no new session.
\item \emph{Labeling of test-informed results.} Round~2 carries the label test-informed. The recommended model and its test error, 0.0437\,m, were the same in both rounds: the interaction changed the content of the pool and the understanding of the error, not the accuracy of the recommended model.
\end{enumerate}

Every question, answer, request and change of the task is part of the session record.

\section{Discussion}
\label{sec:discussion}

\subsection{Effect of the task description}
\label{sec:disc-context}

Prior knowledge did not reduce test error. In all three systems, the mean test score of the described task is higher than that of the blind task (Table~\ref{tab:aspire}): by 0.0029\,m for the mobile robot, by 0.0009\,rad for the friction system and by 0.0008\,bar for the fermenter. We state these differences as observations on five runs each and on small test sets; for the friction system and the fermenter, the difference is smaller than the sample standard deviation of the blind runs (0.0079\,rad and 0.0010\,bar).

For the friction system, the five described scores (0.0292 to 0.0459\,rad) and the five blind scores (0.0322 to 0.0492\,rad) cover nearly the same range. The models differ in form. All five recommended models of the described task contain a mechanism for sticking at rest, which the description and the intended use name, and none of the blind task does; the blind models also violate R2 (Section~\ref{sec:res-approaches}). The test score does not show this difference. The test data contain little of what it concerns: the two free decays, in which the system comes to rest, are 230 of the 978 scored samples.

For the mobile robot, the five blind scores (0.0379 to 0.0415\,m) all lie below the five described scores (0.0417 to 0.0443\,m). A possible explanation is that with the description, the agent recommends the kinematic model with two gains, whereas four of the five blind models contain a dead zone in the response to an input, the term that also lowers the test score in the pools of the described task.

The results of the blind tasks also address the question of recall (Section~\ref{sec:design}). Without any name or description, the agent reached test errors that were not higher, and for two systems, it arrived at the physical structure from the data alone. The low errors, therefore, do not depend on a model that the language model recalls from a description of the system.

The described models consist of terms with a physical name and a unit, and they have fewer parameters for two of the three systems ($2.4 \pm 0.9$ against $9.6 \pm 5.9$ for the mobile robot and $5.6 \pm 1.1$ against $19.0 \pm 2.2$ for the fermenter). For the fermenter, the described task specifies a physically interpretable model, whereas the blind task produces cubic polynomials with comparable test errors. The described models also satisfy physical properties specified in the task requirements: the direction of the gravitational torque at rest for the friction system (R2), and monotonicity and equilibrium conditions for the fermenter (R1–R5; Section~\ref{sec:res-records}). These properties are not reflected in the test error and must be assessed separately or specified as formal requirements. The task description also influences the quality of the model records: in the assessment of Section~\ref{sec:res-records}, 13 of the 18 marks other than \emph{yes} concern blind tasks.

\subsection{Reliability of the model records}
\label{sec:disc-records}

The harness checks the completeness and consistency of model records but cannot verify their content. We therefore assessed the records manually. None of the 30 records contained an undisclosed discrepancy between the declared parameter types and the estimation scripts. However, nine records contained discrepancies between the reported origin of the model structure and the execution logs. In some cases, the record attributed a modeling choice to an analysis that had actually followed it; in others, it described a comparison more broadly than the executed analysis justified. The records, therefore, documented the methods and results more reliably than the reasoning and sequence of decisions that led to them. The execution logs provide the evidence needed to verify these accounts.

\subsection{ASPIRE and a session with a general coding agent}
\label{sec:disc-harness}

A user can give the same data and the same description to a general coding agent in a conversation, with the same language model and the same tools, and ask for a closed-form model. We have no evidence that the harness changes the error of the identified models. ASPIRE provides a curated interface for agentic system identification; it brings value in aspects such as whether a result can be trusted, compared, and reproduced. The test data are outside the agent's reach, and one implementation computes all errors. The recommendation is fixed before any test result exists. The deliverable has a checked form, and the formal requirements are verified on every model. The complete record of a run is stored. A task is repeated with identical settings, so that the variability of the agent is reported with the result. In a conversation, each of these properties depends on what the user arranges and maintains manually.

\subsection{Limitations}
\label{sec:disc-limitations}

The test data are sealed against the agent but not against the authors, who used them during the development of the framework and of the tasks. All runs used one language model, for which the vendor publishes no dated snapshot, so that the results do not establish how ASPIRE performs with other models, and the model behind the identifier may change. While preliminary experiments have shown that ASPIRE also works with different models, it has not been evaluated on a larger scale. Assumptions (A1)--(A3) in Section~\ref{sec:problem} restrict the framework to a single scored output, to systems whose states are measured, and to a given measurement equation. A score that combines several outputs is not part of the framework.

\section{Conclusions}
\label{sec:conclusion}

We proposed ASPIRE, a framework for system identification in which an LLM agent selects and executes the identification method and delivers closed-form models from measured data and the prior knowledge that the user provides. Every model found by ASPIRE is accompanied by a record of its parameters, of their estimation, and of its interpretation; the formal physical requirements are verified on it; and its error is computed on test data that the agent cannot access. ASPIRE runs autonomously or in a dialogue with the user.

We evaluated ASPIRE on three measured systems, each with and without a description of the system. All 30 runs delivered valid models, at a cost of 0.96 to 3.38 US dollars per run. For the rotational motion under dry friction, the task stated formal requirements on the direction of the gravitational torque and on the polarity of the motor, and the verification demonstrated no violation in any model found by the agent. Table~\ref{tab:methods} compares the test errors of the recommended models with those of the models found by PySINDy and PySR. Withholding the description did not increase the test error. It changed the form of the models: with the description, the models consisted of physically named terms and, for two systems, had fewer parameters. In the assessment of the model records of the 30 recommended models, 132 of the 150 marks were \emph{yes}, 18 \emph{partly} and none \emph{no} (Table~\ref{tab:assessment}). The records documented the methods and the results more reliably than the sequence of decisions that led to them: nine records differed from the execution logs in one claim on the origin of the model (Section~\ref{sec:disc-records}). In an interactive session on the data of the mobile robot, an analysis requested by the user attributed 83\,\% of the squared one-step prediction error of the heading to the first sample after a change of the commands, and the user's feedback led to a smaller pool with simpler expressions, while the test error of the recommended model did not change (Section~\ref{sec:res-interactive}).

In our future research, we will extend ASPIRE to systems with several outputs and with unmeasured states, compare several language models, and evaluate the interactive operation with users who did not take part in the development of the framework. We will also evaluate how the agent can guide the collection of data in interactive operation: from the training and validation data and the intended use, the agent suggests to the user which data to obtain, so that the operating regions relevant to the problem that the data do not yet sufficiently cover are covered, and the user adds the new data between the identification rounds. We will also equip the harness with an agent that can assess the compliance of the models with the requirements stated in natural language.

\section*{CRediT authorship contribution statement}

\textbf{Erik Derner:} Conceptualization, Methodology, Software, Investigation, Writing -- original draft. \textbf{Ji\v{r}\'i Kubal\'ik:} Methodology, Data curation, Writing -- review \& editing. \textbf{Robert Babu\v{s}ka:} Methodology, Supervision, Writing -- review \& editing.

\section*{Declaration of competing interest}
The authors declare that they have no known competing financial interests or personal relationships that could have appeared to influence the work reported in this paper.

\section*{Declaration of generative AI and AI-assisted technologies in the manuscript preparation process}
During the preparation of this work the authors used Claude (Anthropic) and ChatGPT (OpenAI) to assist with drafting and editing the manuscript text and with implementing the software described herein, under the authors' direction. The inputs of the user in the interactive session of Section~\ref{sec:res-interactive} were drafted with the same assistance from the notes of the author who acted as the user, and were reviewed and adopted by that author. The writing assistance is distinct from the agentic systems that constitute the object of study, whose runs are reported as experimental results. After using these tools, the authors reviewed and edited the content as needed and take full responsibility for the content of the published article.

\section*{Acknowledgements}

This work was co-funded by the European Union under the project ROBOPROX (reg.\ no.\ CZ.02.01.01/00/22\_008/0004590). This work has also received funding from the European Union's Horizon Europe research and innovation programme under grant agreement No 101070254 (CoreSense). This work was also supported by the European Union's Horizon Europe research and innovation programme under grant agreement No 101214398 (ELLIOT). Funded by the European Union. Views and opinions expressed are however those of the authors only and do not necessarily reflect those of the European Union or the European Commission. Neither the European Union nor the European Commission can be held responsible for them.

\section*{Data availability}
The harness implementation, the task specifications with their training, validation and test data, the experiment configurations, the complete run records of the six experiments and the reports of these experiments as HTML pages are available at \repositorylink{} under the Creative Commons Attribution 4.0 International licence (CC~BY~4.0). The run records contain every request to the language model, every response and tool call, the working directory of the agent with its scripts, the delivered models and their evaluation, together with the software versions and all settings of every run. The repository also contains the scripts, settings and models of the PySINDy and PySR comparisons, the script that computes the reported numbers, tables and figures from the run records, and the assessment of the model records with the evidence for every mark. The test data are sealed against the agent during a run; they are not withheld from readers.

\appendix
\counterwithin{equation}{section}
\counterwithin{table}{section}
\counterwithin{figure}{section}

\section{System prompt}
\label{app:prompt}

The harness supplies the following system prompt in autonomous operation. \texttt{TASK.md} is the task brief (Section~\ref{sec:task}), and \texttt{FORMAT.md} describes the format of the model record and the rules for the agent's texts (Section~\ref{sec:candidates}). The harness replaces \texttt{\{packages\}} by the names of the SR packages that it finds installed in the agent's environment.

\begin{itshape}
You are an agent for system identification. Your task is in TASK.md in the working directory; read it first. FORMAT.md describes the files you deliver and how to write your texts. A human will assess your models and your explanations.

How you identify the models is your decision. You may read and write files in the working directory, write and run code, fetch web resources, search the web, install Python packages and start helper agents. The installed symbolic-regression packages are \{packages\}; symbolic-regression search is suited to tasks of this kind, and whether you use it is your decision. Models of any kind, including black-box models, may serve as intermediate tools; the models you deliver have the form that TASK.md requires.

Report what you actually did: the methods, the data and the settings you used. Helper agents cannot start further helpers, and their turns and cost count toward your limits. Everything you do is recorded. The test data and the evaluation results are not available to you; do not try to reach them or any file outside the working directory.
\end{itshape}

In interactive operation, the following paragraph is appended: \textit{This is an interactive session; TASK.md states how it proceeds. When you ask the user, ask one specific question at a time and say which decision it affects. If the user cannot answer, state the assumption you make instead and its limitation.} The harness adds a description of the environment (the command interpreter and the installed packages with their versions) to the prompt.

\section{Recommended models of the described tasks}
\label{app:records}

This appendix gives the recommended candidates of the three described tasks, each from the run with the median test score of the five runs, as the agent delivered them: the expressions and the parameter table, with values rounded to four significant digits and descriptions quoted and shortened. In the quotations, omissions are marked by [\ldots], and names that the agent wrote as code are set in a typewriter font. The complete model records are in the repository.

\emph{Mobile robot} (test score $S = 0.0428$\,m; training error 0.0406\,m). The agent names the family ``Calibrated unicycle kinematics with forward Euler discretization''. The delivered expressions are
\begin{align}
x(k+1) &= x(k) + T_s\, g_f\, v_f(k) \cos \phi(k), \nonumber\\
y(k+1) &= y(k) + T_s\, g_f\, v_f(k) \sin \phi(k), \label{eq:rec-robot}\\
\phi(k+1) &= \phi(k) + T_s\, g_a\, v_a(k). \nonumber
\end{align}
Table~\ref{tab:rec-robot} is its parameter table.

\begin{table*}[pos=tp]
\centering
\caption{Parameter table of the model (\ref{eq:rec-robot}) of the mobile robot as delivered by the agent; values rounded to four significant digits, descriptions quoted and shortened.}
\label{tab:rec-robot}
\footnotesize
\setlength{\tabcolsep}{4pt}
\begin{tabular}{llllL{0.5\textwidth}}
\toprule
Name & Value & Unit & Type & Description \\
\midrule
$T_s$ & 0.2 & s & fixed & ``Sampling period \texttt{Ts = 0.2} s supplied in TASK.md; used to convert velocity to a one-sample increment.'' \\
$g_f$ & 0.8508 & -- & fitted & ``Forward command gain; coefficient of \texttt{v\_f} in effective forward velocity. [\ldots]'' \\
$g_a$ & 0.7642 & -- & fitted & ``Angular command gain; coefficient of \texttt{v\_a} in effective angular velocity. [\ldots]'' \\
\bottomrule
\end{tabular}
\end{table*}

\emph{Friction system} (test score $S = 0.0442$\,rad; training error 0.0616\,rad). The agent names the family ``Regularized Coulomb-viscous friction with a static dead zone and independent sliding/static speed scales''. With the abbreviation $D = a u - g \sin\theta$, which the record calls the available acceleration, the delivered expressions are
\begin{align}
\frac{\mathrm{d}\theta}{\mathrm{d}t} &= \omega, \nonumber\\
\frac{\mathrm{d}\omega}{\mathrm{d}t} &= D - b\,\omega - c \tanh\!\left(\frac{\omega}{v}\right) \dcbreak - \exp\!\left(-\frac{\omega^2}{v_s^2}\right) \min\!\left(\max(D, -c),\, c\right). \label{eq:rec-friction}
\end{align}
At rest, the last term cancels $D$ whenever $|D| \leq c$, so that the acceleration is zero over an interval of angles and of voltages. Away from rest, the weight $\exp(-\omega^2/v_s^2)$ of this term vanishes, and the sliding friction $c \tanh(\omega/v)$ remains. The harness proved R1 for this model and found no violation of R2 and R3 at the sampled points. The record derives both requirements from the expression; for R2 it states: ``At \texttt{omega = 0} and \texttt{u = 0}, \texttt{D-Min(Max(D,-c),c)} has the same sign as \texttt{D}, or is zero; therefore R2 holds.'' According to the record, the six parameters of Table~\ref{tab:rec-friction} were estimated together on the training and validation data by minimizing the squared errors of the simulated angle, from three sets of starting values.

\begin{table*}[pos=tp]
\centering
\caption{Parameter table of the model (\ref{eq:rec-friction}) of the friction system as delivered by the agent; values rounded to four significant digits, descriptions quoted.}
\label{tab:rec-friction}
\footnotesize
\setlength{\tabcolsep}{4pt}
\begin{tabular}{llllL{0.46\textwidth}}
\toprule
Name & Value & Unit & Type & Description \\
\midrule
$a$ & 28.52 & rad/(s$^2$\,V) & fitted & ``Motor-voltage to acceleration gain \texttt{a}; only gain divided by inertia is identified.'' \\
$g$ & 112.1 & rad/s$^2$ & fitted & ``Positive restoring-acceleration scale \texttt{g}.'' \\
$b$ & 1.591 & 1/s & fitted & ``Positive viscous acceleration coefficient \texttt{b}; includes motor back-emf damping and bearing friction without separating them.'' \\
$c$ & 22.22 & rad/s$^2$ & fitted & ``Dry-friction acceleration scale \texttt{c}.'' \\
$v$ & 0.5181 & rad/s & fitted & ``Sliding-friction regularization speed \texttt{v} in \texttt{tanh(omega/v)}.'' \\
$v_s$ (\texttt{vs}) & 0.1645 & rad/s & fitted & ``Independent static-friction localization speed \texttt{vs} in \texttt{exp(-(omega/vs)**2)}.'' \\
\bottomrule
\end{tabular}
\end{table*}

\emph{Fermenter} (test score $S = 0.0229$\,bar; training error 0.0248\,bar). The agent names the family ``Grey-box nonlinear lumped isothermal gas-inventory model with regularized signed power-law outlet flow and an empirical rational valve characteristic''. With the opening of the valve $s = \min(1, \max(0, 1 - u_\mathrm{v}/v_\mathrm{max}))$ and the relative pressure difference $z = (P - p_\mathrm{atm})/p_\mathrm{atm}$, two abbreviations of ours, the delivered expression is
\begin{equation}
\frac{\mathrm{d}P}{\mathrm{d}t} = a_\mathrm{in}\,u_\mathrm{f} - c_f\, \frac{(1 + \sigma)\, s^{n}}{1 + \sigma s^{n}}\, \frac{z}{\left(z^2 + \delta^2\right)^{(1-q)/2}}.
\label{eq:rec-fermenter}
\end{equation}
Table~\ref{tab:record} is its parameter table, in which we write the parameters as symbols. The description of $\rho$ in the table is explained as follows; the explanation is ours and is not part of the record. The parameter $\rho$ enters (\ref{eq:rec-fermenter}) only through the tied parameter $\delta$. For a fully open valve ($s = 1$) and a pressure close to the atmospheric pressure ($|z| \ll \delta$), the model reduces to $\mathrm{d}P/\mathrm{d}t \approx a_\mathrm{in}\,u_\mathrm{f} - \rho\,(P - p_\mathrm{atm})$, so that $\rho$ is the rate at which the pressure relaxes to the atmospheric pressure. The harness integrates the model by one step of the fourth-order Runge--Kutta method per sampling period $T_s = 5$\,s (Section~\ref{sec:eval}). For a linear relaxation with the rate $\rho$, this integration is stable only for $\rho T_s < 2.785$, that is, for $\rho < 0.557$\,s$^{-1}$. The agent bounded $\rho$ by 0.5\,s$^{-1}$, and the delivered value equals this bound.

\begin{table*}[pos=tp]
\centering
\caption{Parameter table of the model (\ref{eq:rec-fermenter}) of the fermenter as delivered by the agent; values rounded to four significant digits, descriptions quoted and shortened. The names of the agent are given in parentheses. The tied parameter $\delta$ is computed by the equation $\delta = (c_f/(p_\mathrm{atm}\,\rho))^{1/(1-q)}$, which the record states. The description of $\rho$ is explained in the text.}
\label{tab:record}
\footnotesize
\setlength{\tabcolsep}{4pt}
\begin{tabular}{llllL{0.46\textwidth}}
\toprule
Name & Value & Unit & Type & Description \\
\midrule
$a_\mathrm{in}$ (\texttt{ain}) & $1.230 \times 10^{-3}$ & bar\,min/(s\,L) & fitted & ``Inlet conversion gain \texttt{ain}: pressure rate per measured \texttt{flow} [\ldots]'' \\
$c_f$ (\texttt{cf}) & 0.1108 & bar/s & fitted & ``Full-open outlet scale \texttt{cf} in bar/s for the dimensionless pressure-driving law. [\ldots]'' \\
$n$ (\texttt{expo}) & 1.364 & -- & fitted & ``Positive aperture exponent \texttt{expo} [\ldots]. It describes the unknown actuator-to-effective-opening characteristic; it is not a measured valve geometry.'' \\
$\sigma$ (\texttt{shape}) & 2.539 & -- & fitted & ``Nonnegative series-loss shape \texttt{shape} [\ldots]. The rational aperture law saturates as the opening grows [\ldots]'' \\
$\rho$ (\texttt{rate}) & 0.5000 & 1/s & fitted & ``Full-open atmospheric small-signal relaxation rate \texttt{rate} [\ldots]. The upper bound of \texttt{rate = 0.5} 1/s was chosen for stable five-second RK4 relaxation [\ldots]; the final estimate reaches that bound.'' \\
$q$ (\texttt{power}) & 0.5409 & -- & fitted & ``Far-from-atmosphere pressure exponent \texttt{power} [\ldots]. It interpolates empirical fractional pressure-loss behavior and is not assumed to equal an isentropic exponent.'' \\
$p_\mathrm{atm}$ (\texttt{patm}) & 1 & bar & fixed & ``Atmospheric reference \texttt{patm = 1} bar, taken from the approximately one-bar equilibrium in TASK.md and deliberately not estimated; it fixes the equilibrium rather than absorbing pressure-sensor bias.'' \\
$v_\mathrm{max}$ (\texttt{vmax}) & 100 & \% & fixed & ``Fully closed calibration \texttt{vmax = 100} percent, given in TASK.md [\ldots]'' \\
$\delta$ (\texttt{delta}) & 0.03755 & -- & tied & ``Dimensionless near-atmospheric pressure scale \texttt{delta}; computed from the outlet scale and the fitted atmospheric relaxation rate. This gives a finite near-equilibrium slope rather than a singular fractional-power slope.'' \\
\bottomrule
\end{tabular}
\end{table*}

\section{Physical requirements of the fermenter}
\label{app:requirements}

This appendix gives the physical requirements of the described task of the fermenter. The statements R1 to R5 are quoted in full, as the agent received them.
\begin{itemize}
\item[R1] ``Monotonicity with respect to inlet flow. At a fixed pressure and valve position, increasing the inlet air flow rate cannot decrease the pressure derivative.''
\item[R2] ``Monotonicity with respect to valve closure. At a fixed pressure and inlet flow rate, closing the outlet valve further cannot decrease the pressure derivative.''
\item[R3] ``Zero-flow equilibrium. With zero inlet flow and an open outlet valve, atmospheric pressure (approximately 1 bar) is the pressure equilibrium.''
\item[R4] ``Pressure relaxation. With zero inlet flow and an open outlet valve, pressure tends toward atmospheric pressure.''
\item[R5] ``Fully closed valve. For a perfectly sealed tank with a fully closed outlet valve, the pressure cannot decrease when air is supplied.''
\end{itemize}

\bibliographystyle{cas-model2-names}
\bibliography{refs}

\end{document}